\documentclass{article}

\PassOptionsToPackage{numbers,square}{natbib}
\usepackage[eandd,final]{neurips_2026}

\usepackage[utf8]{inputenc} %
\usepackage[T1]{fontenc} %
\usepackage{hyperref} %
\usepackage{url} %
\usepackage{booktabs} %
\usepackage{amsfonts} %
\usepackage{nicefrac} %
\usepackage{microtype} %

\usepackage[dvipsnames, table]{xcolor}

\usepackage{graphicx}
\usepackage{amsmath}
\usepackage{amssymb}
\usepackage{booktabs}
\usepackage{mathtools}
\usepackage{iftex}
\ifPDFTeX
  \usepackage[accsupp]{axessibility}  %
\fi
\usepackage{color}
\usepackage{array}
\usepackage{pdflscape}
\usepackage[longtable]{multirow}
\usepackage{longtable}
\usepackage{tabularray}
\usepackage{booktabs}
\usepackage{times}
\usepackage{epsfig}
\usepackage{amsmath}
\usepackage{bbm}
\DeclareMathAlphabet\mathbfcal{OMS}{cmsy}{b}{n}
\usepackage{makecell}
\usepackage{siunitx}
\usepackage{svg}
\usepackage{float}
\usepackage{wrapfig}
\usepackage{needspace}
\usepackage{comment}
\usepackage{lipsum}
\usepackage{stfloats}
\usepackage{multicol}
\usepackage{multirow}
\usepackage{bm}
\usepackage{etoolbox}
\usepackage{icomma}
\usepackage{array}
\usepackage{tabulary}
\usepackage{xcolor}
\usepackage{paralist}
\usepackage{booktabs}
\usepackage{adjustbox}
\usepackage{caption}
\usepackage{subcaption}
\usepackage{arydshln}
\usepackage{rotating}
\usepackage{enumitem}
\ifPDFTeX
  \usepackage{fontawesome5}
\else
  \usepackage{fontawesome}
\fi

\definecolor{gray}{rgb}{0.3,0.3,0.3}
\definecolor{blue}{rgb}{0,0.5,1}
\definecolor{mask_red}{rgb}{1,0,0.8}
\definecolor{green}{rgb}{0.2,1,0.2}
\definecolor{rblue}{rgb}{0,0,1}
\definecolor{lightblue}{HTML}{6495ed}
\definecolor{lightred}{HTML}{F19C99}

\definecolor{graytablerow}{gray}{0.6}

\usepackage{pifont}%
\usepackage{tabu}

\usepackage{tikz}

\usepackage[capitalize]{cleveref}
\crefname{section}{Sec.}{Secs.}
\Crefname{section}{Section}{Sections}
\Crefname{table}{Table}{Tables}
\crefname{table}{Tab.}{Tabs.}

\definecolor{abstracturl}{HTML}{D2508C}

\title{%
  \texorpdfstring{%
    \begin{tabular}{@{}c@{\hspace{0.5em}}c@{}}
      \begin{minipage}[c]{2.4em}
        \includegraphics[width=\linewidth]{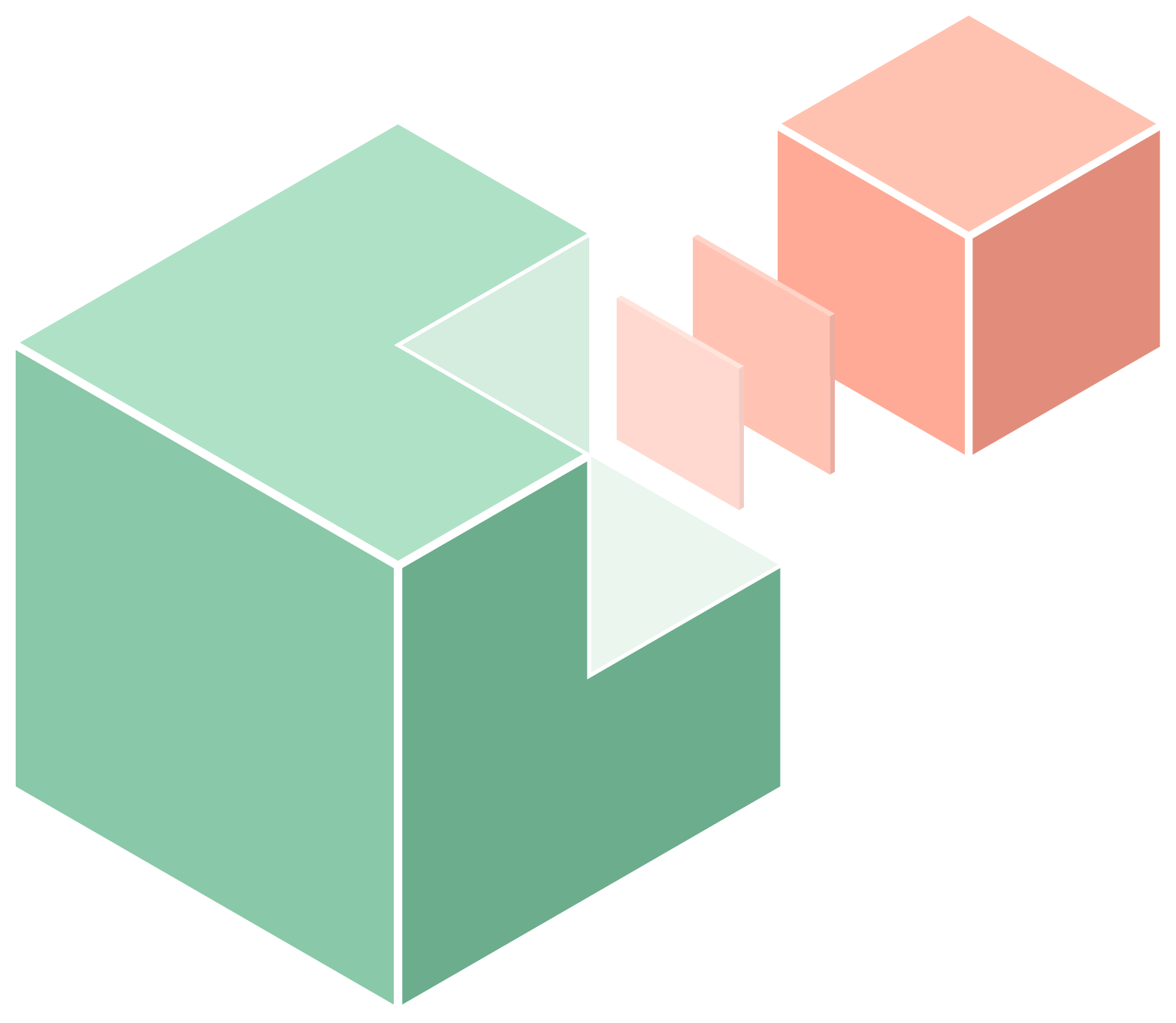}%
      \end{minipage}
      &
      \begin{tabular}[c]{@{}c@{}}
        OccStress: \underline{Stress}-Testing the \\
        4D \underline{Occ}upancy Forecasting Chain
      \end{tabular}
    \end{tabular}%
  }{OccStress: Stress-Testing the 4D Occupancy Forecasting Chain}%
}

\author{%
  Yu Zheng\textsuperscript{1}\quad
  Jie Hu\textsuperscript{1}\quad
  Jiaqi Xiong\textsuperscript{2}\quad
  Ruiping Liu\textsuperscript{3}\\[0.4em]
  \textbf{Junwei Zheng}\textsuperscript{3,4}\quad
  \textbf{Kailun Yang}\textsuperscript{1}\quad
  \textbf{Jiaming Zhang}\textsuperscript{1,}\thanks{Corresponding author, (\href{mailto:jiamingzhang@hnu.edu.cn}{jiamingzhang@hnu.edu.cn}).}\\[0.7em]
  {\small\textsuperscript{1}Hunan University}\\[0.2em]
  {\small\textsuperscript{2}University of Oxford}\\[0.2em]
  {\small\textsuperscript{3}Karlsruhe Institute of Technology}\\[0.2em]
  {\small\textsuperscript{4}ETH Zurich}
}

\begin{document}
  \maketitle

  \begin{figure}[H]
    \centering
    \includegraphics[width=1.0\linewidth]{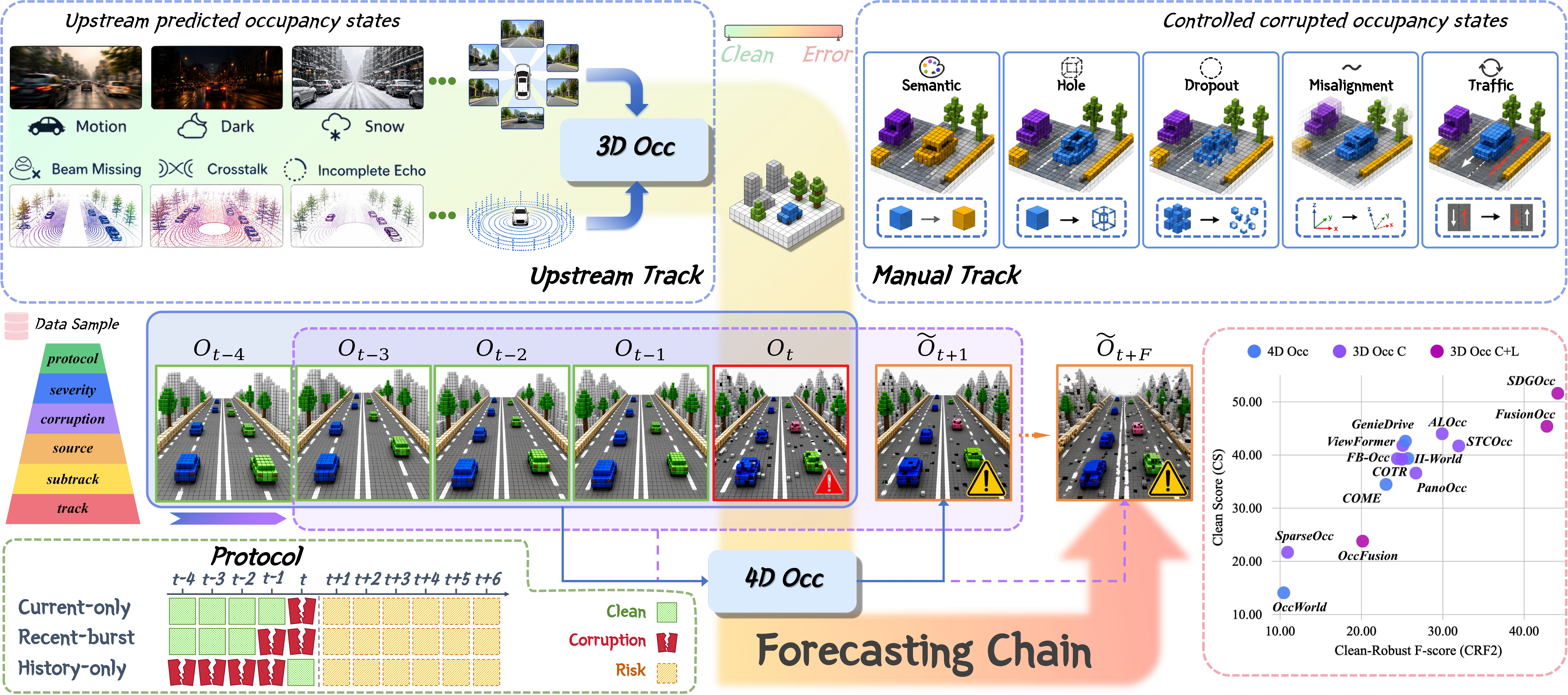} 
    \caption{Overview of the 4D occupancy forecasting chain studied in \textbf{OccStress}.
    Occupancy world models rely on temporal occupancy states for future
    forecasting, while errors from upstream perception or direct state
    perturbations can persist and propagate across the forecasting horizon.}
    \label{fig:teaser}
  \end{figure}

  \begin{abstract}
    Occupancy world models use historical occupancy states to forecast future 3D scenes, but their robustness under corrupted temporal inputs remains poorly understood. Existing evaluations primarily emphasize clean forecasting accuracy and provide limited evidence about how errors enter, persist, and propagate through the occupancy perception-forecasting chain. This paper introduces \textbf{OccStress, a robustness stress-testing benchmark for the occupancy forecasting chain}. OccStress contains \textbf{21 corruption families with 61 severity configurations and 10,827 strict temporal anchors across 3 datasets}. OccStress covers both 3D occupancy perception and 4D occupancy forecasting through two complementary tracks. This design separates model-mediated pipeline errors under standardized sensor stressors from the intrinsic sensitivity of 4D forecasting models to corrupted occupancy states. OccStress further defines temporal injection protocols to test whether errors in the current state, recent history, or earlier history affect future forecasts differently. OccStress provides aggregate metrics for evaluating robustness along the occupancy forecasting chain. Experiments with five 4D forecasters reveal that current occupancy models are substantially affected by both upstream prediction errors and direct state corruptions, and that clean performance alone is an insufficient description of source- and position-specific robustness. Code, data, and evaluation tools are available at 
    {\hypersetup{pdfborder={0 0 0}}%
    \href{https://insailab.org/OccStress}{%
      \textcolor{abstracturl}{%
        \nolinkurl{https://insailab.org/OccStress}%
      }%
    }}.
  \end{abstract}

  \section{Introduction}

3D semantic occupancy has become a central representation for autonomous driving, as it provides dense voxel-level geometry and semantics for downstream prediction and planning. Recent work has extended occupancy from single-frame perception to 4D occupancy forecasting, where historical occupancy states are encoded, fused, and rolled out to predict future 3D scenes. This shift makes occupancy not only a perception output, but also an intermediate state in a longer forecasting chain.

Current evaluation protocols, however, largely assume that this chain is clean. Observations are treated as reliable, upstream occupancy states are assumed to be accurate, and historical context is assumed to be consistent. These assumptions are fragile in deployment. Sensor degradations may be converted into erroneous 3D occupancy predictions; upstream state errors may be stored in the temporal context; and corrupted historical states may be encoded, fused, and propagated into future forecasts. Robustness in this setting is therefore not simply a property of a single frame, but a property of the full occupancy forecasting chain.

Existing benchmarks~\cite{xie2025benchmarking,kong2023robo3d,tian2023occ3d} only partially address this issue. Occupancy benchmarks mainly evaluate clean perception or clean forecasting, while robustness benchmarks in autonomous driving mostly stress-test instantaneous perception tasks under camera, LiDAR, or multi-sensor corruptions. They provide limited evidence about how errors enter the temporal occupancy state, whether they persist after the current input recovers, and how they affect long-horizon forecasts. This leaves a gap between perception robustness and 4D occupancy forecasting robustness.

We introduce \textbf{OccStress}, an evaluation benchmark for stress-testing the 4D occupancy forecasting chain. As shown in Figure~\ref{fig:teaser}, OccStress separates two complementary error pathways. (1) In the \textit{upstream} track, corrupted sensor observations are first processed by 3D occupancy estimators, and the resulting predicted occupancy states are fed to downstream 4D forecasting models. This track measures model-mediated pipeline errors under standardized sensor stressors. (2) In the \textit{manual} track, clean occupancy states are directly perturbed before forecasting, exposing the intrinsic sensitivity of 4D models to semantic, geometric, alignment, and traffic-layout state corruptions.

For the first time, OccStress includes three temporal challenges. OccStress asks \textit{where} the error enters the temporal context. (1) \textit{Current-only} corrupts the latest state while keeping history clean, testing whether history can stabilize a degraded current input. (2) \textit{Recent-burst} corrupts temporally adjacent recent states, representing short failures near the forecasting anchor. (3) \textit{History-only} corrupts historical states while keeping the current state clean, testing whether corrupted history can still bias future predictions after apparent recovery.

To summarize, our \textbf{contributions} (\textcolor[RGB]{39,105,182}{\textbf{C}}) are as follows.
\begin{itemize}[leftmargin=0pt,label={},labelsep=0pt,itemindent=0pt,itemsep=1pt,topsep=0pt,parsep=0pt,partopsep=0pt]
  \item \textcolor[RGB]{39,105,182}{\textbf{C1}} \textbf{Chain-level formulation.} We cast 4D occupancy robustness as a forecasting chain problem, tracing errors from observations or states to future predictions.
  \item \textcolor[RGB]{39,105,182}{\textbf{C2}} \textbf{OccStress benchmark.} OccStress stress-tests two error paths across OccStress-nuScenes, OccStress-Waymo, and OccStress-CARLA, containing 21 corruption families with 61 severity configurations and 10,827 temporal anchors.
  \item \textcolor[RGB]{39,105,182}{\textbf{C3}} \textbf{Temporal protocols.} OccStress introduces temporal injection protocols for evaluating robustness in 4D occupancy forecasting chains. We reasonably design three representative protocols.
  \item \textcolor[RGB]{39,105,182}{\textbf{C4}} \textbf{Diagnostic robustness aggregation.} We provide source-wise, protocol-wise, and aggregate scores that decompose robustness by error source and temporal injection position.
  \item \textcolor[RGB]{39,105,182}{\textbf{C5}} \textbf{Reusable evaluation assets.} We release protocol records, corruption generators, evaluation scripts, and benchmark metadata for reproducible stress-testing of occupancy perception and forecasting.
\end{itemize}

By evaluating the forecasting chain with OccStress, we summarize the following \textbf{findings} (\textcolor[RGB]{174,74,63}{\textbf{F}}).
\begin{itemize}[leftmargin=0pt,label={},labelsep=0pt,itemindent=0pt,itemsep=1pt,topsep=0pt,parsep=0pt,partopsep=0pt]
  \item \textcolor[RGB]{174,74,63}{\textbf{F1}} \textbf{Clean accuracy is insufficient for robustness.} Clean performance alone does not reveal sensitivity to different error sources and temporal injection positions.
  \item \textcolor[RGB]{174,74,63}{\textbf{F2}} \textbf{Upstream state quality is a bottleneck.} Even clean upstream occupancy predictions recover only part of the clean occupancy-state reference, making state quality a key factor in the forecasting chain.
  \item \textcolor[RGB]{174,74,63}{\textbf{F3}} \textbf{Temporal injection position matters.} Single-state sweeps reveal model-specific sensitivity to the position of an otherwise identical corruption.
  \item \textcolor[RGB]{174,74,63}{\textbf{F4}} \textbf{Temporal fusion design requires more careful evaluation.} Existing methods differ substantially in how they exploit temporal information. The deliberate design specifically for this is still insufficient.
\end{itemize}

  \section{Related Work}

\noindent
\textbf{Occupancy Perception}.
Recent occupancy perception methods incorporate temporal fusion and streaming prediction~\cite{ye2024cvtocc, Zuo2025gaussianworld}. Existing occupancy perception methods can be organized according to their test-time input modalities. \emph{Camera-only} methods predict 3D semantic occupancy from multi-view images by lifting image features into BEV or volumetric representations, often with BEV-based transformers or structured voxel decoders \cite{huang2021bevdet,li2024bevformer,huang2023tri,zhang2023occformer,tian2023occ3d,chen2025alocc,liao2025stcocc,ma2024cotr}.
\emph{Camera--LiDAR} or \emph{camera--point cloud} methods further incorporate point-based geometric cues and usually achieve stronger occupancy estimation through explicit multi-modal fusion \cite{wang2023openoccupancy,zhang2024occfusion,zhang2024fusionocc,duan2025sdgocc}. Benchmark efforts such as Occ3D~\cite{tian2023occ3d} and OpenOccupancy~\cite{wang2023openoccupancy} have promoted both settings and made occupancy prediction a standard task for autonomous driving scene understanding. Despite rapid progress in clean-setting accuracy, existing work mainly emphasizes in-distribution perception quality, while robustness under corrupted observations or corrupted occupancy inputs remains much less explored.

\noindent
\textbf{Occupancy World Models}.
Recent work extends occupancy prediction from single-frame perception to sequential forecasting, treating occupancy as a compact state space for modeling scene evolution over time \cite{ma2024cam4docc,zheng2024occworld,diehl2025dio,chen2025occprophet,li2025semi}. A common trend is to build latent or tokenized occupancy representations together with temporal fusion over historical states, so as to make long-horizon forecasting computationally feasible while preserving semantic and geometric fidelity \cite{zheng2024occworld,liao2025i2,yang2026geniedrive}. More recent studies further introduce controllable or scene-centric formulations to better support planning and decision-making \cite{yang2025driving,shicome}. Despite these advances, existing occupancy world models are still primarily optimized for clean forecasting quality, leaving the robustness of temporal tokenization and history fusion under corrupted inputs largely unexplored.

\begin{wrapfigure}[17]{r}{0.48\textwidth}
  \vspace{-1.5em}
  \centering
  \includegraphics[width=\linewidth]{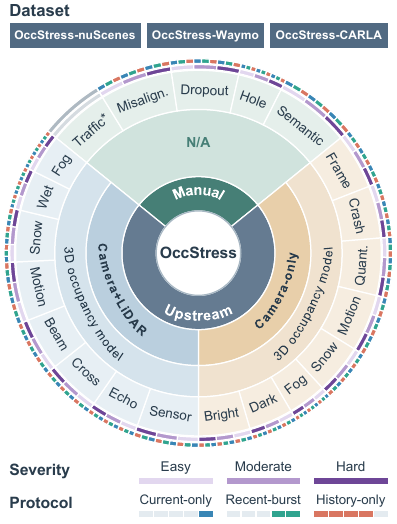}
  \captionsetup{font=small,justification=raggedright,singlelinecheck=false}
  \caption{\textbf{OccStress data distribution}.}
  \label{fig:benchmark_pipeline}
\end{wrapfigure}

\noindent
\textbf{Perception Robustness in Autonomous Driving}.
Robustness benchmarking has been widely studied in autonomous driving perception, with prior work systematically evaluating camera corruptions, LiDAR degradations, and multi-sensor perturbations for tasks such as BEV perception, 3D detection, and map construction \cite{xie2025benchmarking,kong2023robo3d,hao2024your,hao2025msc,dong2023benchmarking,wang2025uniocc}. These studies show that standard clean-set rankings often fail to reflect model behavior in adverse conditions, and establish corruption-based evaluation as an important complement to conventional accuracy benchmarks. However, existing benchmarks are still centered on instantaneous perception. In contrast, robustness for occupancy world models is inherently temporal. Our work explicitly targets temporal robustness in 4D occupancy forecasting.

  \section{OccStress Dataset and Benchmark} %

\subsection{Data Construction}

\begin{wrapfigure}[3]{r}{0.48\textwidth}
\end{wrapfigure}

\textbf{OccStress}, a robustness stress-testing benchmark for occupancy forecasting chain, covers both 3D occupancy perception and
4D occupancy forecasting. OccStress contains 21 corruption families with 61 severity configurations and 10,827 strict H4/F6 anchors across OccStress-nuScenes, OccStress-Waymo, and OccStress-CARLA. In 4D occupancy forecasting,
a corrupted occupancy state can be encoded by the tokenizer, fused with
historical context, and carried forward into multiple future predictions.
Robustness depends not only on the severity of the input degradation,
but also on how the forecasting model uses, suppresses, or amplifies corrupted
history. Figure~\ref{fig:benchmark_pipeline} shows the data construction.

OccStress separates two complementary error pathways. In the \textit{upstream}
track, corrupted sensor observations are first processed by a 3D occupancy
estimator, and the resulting occupancy predictions are then used as temporal
inputs to a 4D model. This deployment-style track measures how
observation-level degradations are transmitted through an intermediate
occupancy representation. In the \textit{manual} track, clean occupancy states
are directly perturbed before forecasting. This controlled track exposes the
4D model's sensitivity to state-level semantic, geometric, alignment, and
traffic-layout errors. Together, the two tracks distinguish robustness
inherited from upstream perception from robustness intrinsic to temporal
occupancy forecasting.

Each OccStress evaluation configuration is defined by seven fields:
\[
  (\mathrm{dataset},\ \mathrm{track},\ \mathrm{subtrack},\ \mathrm{source},\
  \mathrm{corruption},\ \mathrm{severity},\ \mathrm{protocol}).
\]
The \emph{dataset} field identifies the OccStress subset. The \emph{track} field selects between upstream pipeline corruption and manual
state corruption. The \emph{subtrack} and \emph{source} fields specify the
sensor branch and the upstream 3D occupancy model when applicable, and are empty
for the manual track. The remaining fields define the \emph{corruption} family, its
\emph{severity} level, and its temporal injection \emph{protocol}. The following subsections
detail these components.

\subsection{Task Formulation}

Let \(\mathbf{O}_{t}\in\{0,\ldots,C\}^{X\times Y\times Z}\) denote the
semantic occupancy state at time \(t\), where each voxel is assigned either a
semantic class or free space. Given \(H\) historical states and the current
state, a 4D occupancy model receives
\begin{equation}
  \mathbf{X}_{t} =
  \{\mathbf{O}_{t-H}, \ldots, \mathbf{O}_{t-1}, \mathbf{O}_{t}\},
\end{equation}
and predicts a sequence of \(F\) future occupancy states,
\begin{equation}
  \hat{\mathbf{Y}}_{t+1:t+F}
  = f_{\theta}(\tilde{\mathbf{X}}_{t}),
\end{equation}
where \(\tilde{\mathbf{X}}_{t}\) denotes the possibly corrupted temporal input.
For ordinary robustness evaluation, the target sequence
\(\mathbf{Y}_{t+1:t+F}\) remains clean, so that the measured degradation can be
attributed to corrupted inputs rather than to a change in supervision. The only
exception is traffic, which simulates changes between left-hand and
right-hand traffic conventions across regions. The same left-right
transformation is therefore applied to both inputs and targets to maintain a
consistent traffic layout.

The three subsets are built from Occ3D-nuScenes~\cite{tian2023occ3d,caesar2020nuscenes}, Occ3D-Waymo~\cite{tian2023occ3d,sun2020scalability}, and UniOcc-CARLA~\cite{wang2025uniocc}, respectively, with \(H{=}4\) historical frames and \(F{=}6\) future frames at 2 Hz. Each frame-level occupancy state
is stored once, while each evaluation sample is described by a protocol record
specifying the history, current input, target sequence, corruption metadata,
and source track. This representation keeps the clean target policy explicit
and allows clean, upstream, and manual settings to share the same temporal
backbone. This window defines anchor eligibility, while each model retains its native input length (Appendix~\ref{sec:app_transfer_setup}).

\subsection{Benchmark Tracks}

\subsubsection{Upstream Pipeline Track}

The \textit{upstream} track evaluates robustness under errors that arise before
4D forecasting. Given a 3D occupancy estimator \(g_{\phi}\), clean or corrupted
sensor observations are converted into occupancy predictions,
\begin{equation}
  \tilde{\mathbf{O}}_{t} = g_{\phi}(\mathcal{A}(\mathbf{S}_{t})),
\end{equation}
where \(\mathbf{S}_{t}\) denotes the sensor observation and \(\mathcal{A}\)
denotes an observation corruption operator. These predicted occupancy states
form the temporal input source for the downstream 4D model, while future
supervision remains the clean occupancy sequence.

Upstream evaluation is organized by sensor modality and source model. OccStress
supports a \textit{Camera-only} subtrack and a
\textit{Camera+LiDAR} subtrack. The camera and point-cloud corruptions 
settings are shown in Table~\ref{tab:1_corruption_severity_level}. For each modality branch, candidate 3D occupancy
models are first evaluated under observation corruption, and a robust
representative source model is then selected for downstream 4D forecasting
experiments.

Because upstream source models can differ substantially even under clean
observations, the clean reference is defined separately for each source. The
corrupted 4D scores are compared against the clean predictions produced by the
same upstream model, rather than against a shared oracle reference. This
source-conditioned reporting avoids conflating upstream occupancy quality with
the temporal robustness of the 4D forecasting model. Unperturbed input positions use the same upstream model's clean predictions, while future targets remain ground truth.

\subsubsection{Manual State-Corruption Track}

The \textit{manual} track directly corrupts clean occupancy states and removes
the upstream perception model from the evaluation loop. This track is designed
to expose when given an erroneous
state in its input context, the model may use history to compensate for the
error, or it may propagate the corrupted state into future predictions.

The manual track introduces five occupancy-state corruption families.
\textit{semantic} corruption changes semantic labels while preserving occupied
support, probing sensitivity to state-level semantic errors. \textit{hole} corruption removes
interior object voxels and creates incomplete or hollow structures.
\textit{dropout} corruption removes occupied voxels within local spatial
regions, emulating missing occupancy evidence. \textit{misalignment}
corruption perturbs temporal consistency through pose-level transformations
rather than direct tensor editing. \textit{traffic} corruption mirrors the
scene left and right in ego coordinates, simulating a structured shift between
left-hand and right-hand traffic conventions. This setting is motivated by the
fact that driving direction and lane organization vary across countries and
regions. The severity levels for all corruption families are summarized in
Table~\ref{tab:1_corruption_severity_level}.

\begin{table*}[t]
    \centering
    \small
    \setlength{\tabcolsep}{5pt}
    \caption{\textbf{Severity level setups in corruption simulations of OccStress}. We consider 2D image,
    3D point cloud, and 4D occupancy-state corruptions. The image corruption settings
    are adapted from RoboBEV~\cite{xie2025benchmarking}, the point cloud
    corruption settings follow Robo3D~\cite{kong2023robo3d}, and the occupancy-state
    corruptions are newly introduced in OccStress.}
    \label{tab:1_corruption_severity_level}

    \begin{adjustbox}
        {max width=\textwidth}
        \begin{tabular}{c c c c c}
            \toprule \textbf{Corruption}                                     & \textbf{Parameter}                  & \textbf{Easy}                    & \textbf{Moderate}              & \textbf{Hard}                  \\
            \midrule

\rowcolor{gray!10}\multicolumn{5}{c}{\textbf{Image Corruptions}}          \\
            \midrule Bright                                                  & adjustment in HSV space             & 0.2                              & 0.4                            & 0.5                            \\
            Dark                                                             & scale factor                        & 0.5                              & 0.4                            & 0.3                            \\
            Fog                                                              & (thickness, smoothness)             & (2.0, 2.0)                       & (2.5, 1.5)                     & (3.0, 1.4)                     \\
            \multicolumn{1}{c}{\multirow{2}{*}{Snow}}                        & (mean, std, zoom, threshold,        & (0.1, 0.3, 3.0, 0.5,             & (0.2, 0.3, 2, 0.5,             & (0.55, 0.3, 4, 0.9,            \\
                                                                             & blur radius, blur std, blending ratio) & 10.0, 4.0, 0.8)                & 12, 4, 0.7)                    & 12, 8, 0.7)                    \\
            Motion                                                           & (radius, sigma)                     & (15, 5)                          & (15, 12)                       & (20, 15)                       \\
            Quant                                                            & bit number                          & 5                                & 4                              & 3                              \\
            Crash                                                            & camera-index draws (with replacement) & 2                              & 4                              & 5                              \\
            Frame                                                            & frame dropping probability          & 2 / 6                            & 4 / 6                          & 5 / 6                          \\
            \midrule \rowcolor{gray!10}\multicolumn{5}{c}{\textbf{Point Cloud Corruptions}}     \\
            \midrule Fog                                                     & beta                                & 0.008                            & 0.05                           & 0.2                            \\
            Wet                                                              & (water height, noise floor)         & (0.2, 0.2)                       & (1.0, 0.3)                     & (1.2, 0.7)                     \\
            Snow                                                             & (snowfall rate, terminal velocity)  & (0.5, 2.0)                       & (1.0, 1.6)                     & (2.5, 1.6)                     \\
            Motion                                                           & trans std                           & 0.2                              & 0.3                            & 0.4                            \\
            Beam                                                             & beam number to drop                 & 8                                & 16                             & 24                             \\
            Cross                                                            & percentage                          & 0.03                             & 0.07                           & 0.12                           \\
            Echo                                                             & drop ratio                          & 0.75                             & 0.85                           & 0.95                           \\
            Sensor                                                           & beam number to drop                 & 8                                & 16                             & 20                             \\
            \midrule \rowcolor{gray!10}\multicolumn{5}{c}{\textbf{Occupancy-State Corruptions}} \\
            \midrule Semantic                                                & (corrupted dynamic, static)         & (1--2, 1--2)                     & (2--3, 2)                      & (3--5, 2--3)                   \\
            Hole                                                             & (instances, removed interior ratio) & (1--2, 0.58--0.78)               & (2--3, 0.78--0.94)             & (4--6, 0.94--1.00)             \\
            Dropout                                                          & (regions, shape, removed fraction)  & (1--2, cuboid, --)               & (3--4, irregular, 0.70--0.90)  & (5--8, irregular, 0.90--1.00)  \\
            Misalignment                                                     & (mode, max translation / yaw)       & (constant, 0.8\,m / 4$^{\circ}$) & (drift, 2.4\,m / 12$^{\circ}$) & (drift, 4.5\,m / 20$^{\circ}$) \\
            Traffic                                                          & left-right mirroring                & -                                & -                              & -                              \\
            \bottomrule
        \end{tabular}
    \end{adjustbox}
\end{table*}

All manual corruption results use the clean ground-truth occupancy sequence as
their reference source. Consequently, the manual track measures sensitivity to
controlled state perturbations, while the upstream track measures degradation
introduced by a full observation-to-state pipeline. This separation allows
OccStress to attribute robustness failures either to upstream perception or to
the temporal forecasting model itself.

\subsection{Temporal Corruption Protocols}

A setting in OccStress is determined not only by the corruption family and
severity, but also by when the corruption enters the temporal context. The same
spatial or semantic disturbance can have different consequences depending on
whether it affects the current state, the most recent history, or the entire
historical window.

\textbf{OccStress uses three temporal injection protocols}. (1) \textit{Current-only}
corrupts only the current input state and keeps all historical states clean. It
tests whether history can stabilize a degraded latest observation.
(2) \textit{Recent-burst} corrupts the current state together with the nearest 1
historical state, representing a short burst of temporally adjacent state
errors. It evaluates sensitivity to corruptions close to the forecasting anchor.
(3) \textit{History-only} corrupts the historical window while keeping the current
input clean for standard state-corruption settings. It tests whether corrupted
history can still bias forecasting after the latest state has recovered.

These protocols jointly vary position, duration, and the number of corrupted
states. The single-state sweep in Figure~\ref{fig:position_sweep} instead varies
only position at a fixed corruption budget.

For \textit{misalignment}, the protocol specifies which frames receive
pose-level perturbations. For \textit{traffic}, it uses a fixed mirrored-sequence
protocol. This special case evaluates structured traffic-layout inversion
rather than noisy input degradation under unchanged labels.

\subsection{Evaluation Metrics}
\label{sec:evaluation_metrics}

Unless otherwise stated, 4D semantic mIoU scores are averaged over the 1s, 2s, and 3s forecasting horizons.
We use \(S\in\{I,P,O\}\) for image, point-cloud, and occupancy-state sources, and \(p\in\{\mathrm{Cur},\mathrm{Burst},\mathrm{Hist}\}\) for temporal protocols. 
\(CS^S\) denotes source-conditioned clean performance, \(RS^S_{\forall}\) averages all corrupted settings for source \(S\), and \(RS^\forall_p\) averages all sources under protocol \(p\). 
Traffic is reported separately and excluded from these averages because it mirrors both inputs and targets.
The summary scores are
\begin{equation}
mRS=\tfrac{1}{3}\sum_S RS^S_{\forall},\quad
CRF_\beta=\tfrac{(1+\beta^2)CS^O\,mRS}{\beta^2CS^O+mRS},\quad
UCR=\tfrac{CS^I+CS^P}{2CS^O}.
\end{equation}
\(mRS\) is the primary aggregate robustness score, \(CRF_\beta\) is a clean-robust balanced score, and \(UCR\) measures upstream clean-state quality. \(CRF_\beta\) is reported on the same mIoU scale as \(CS^O\) and \(mRS\). We report $CRF_2$ by setting $\beta = 2$. Since $\beta>1$, $CRF_\beta$ places greater weight on the robust score.

  \section{Benchmark Evaluation}

\definecolor{avgshade}{gray}{0.93}

\begin{table*}[t]
    \centering
    \small
    \setlength{\tabcolsep}{1.2pt}
    \renewcommand{\arraystretch}{1.14}
    \caption{\textbf{3D occupancy robustness results} on OccStress-nuScenes,
    OccStress-Waymo, and OccStress-CARLA.
    \(\mathrm{C}\) and \(\mathrm{C{+}L}\) denote camera-only and camera--LiDAR
    inputs. \(CS\) is clean semantic mIoU excluding the free class.
    \(RS^{I}_{3D}\) and \(RS^{P}_{3D}\) are corruption-family averaged semantic
    mIoU under image and point-cloud corruptions. For OccStress-CARLA, semantic
    mIoU averages the five occupied validation classes. Bold and underlined
    entries denote the best and second-best results within each dataset and
    input-modality block. ``--'' denotes not applicable.}
    \label{tab:2_3d_occ_robustness}
    \begin{adjustbox}{max width=\textwidth}
    \begin{tabular}{@{}c@{\hspace{2pt}}l c cccccccc c>{\columncolor{avgshade}} c cccccccc c>{\columncolor{avgshade}} c@{}}
        \toprule
        \multirow{2}{*}{\textbf{Input}} & \multirow{2}{*}{\textbf{Method}}
            & \multirow{2}{*}{\textbf{CS}}
            & \multicolumn{8}{c}{\textbf{2D Image Corruptions}}
            & \multirow{2}{*}{$\mathbf{RS}^{\mathbf{I}}_{\mathbf{3D}}$}
            & \multirow{2}{*}{$\mathbf{CRF}^{\mathbf{I}}_{\mathbf{2}}$}
            & \multicolumn{8}{c}{\textbf{3D Point Cloud Corruptions}}
            & \multirow{2}{*}{$\mathbf{RS}^{\mathbf{P}}_{\mathbf{3D}}$}
            & \multirow{2}{*}{$\mathbf{CRF}^{\mathbf{P}}_{\mathbf{2}}$} \\
        \cmidrule(lr){4-11} \cmidrule(lr){14-21}
            & & & \rotatebox{90}{Bright} & \rotatebox{90}{Dark}
              & \rotatebox{90}{Fog}    & \rotatebox{90}{Snow}
              & \rotatebox{90}{Motion} & \rotatebox{90}{Quant}
              & \rotatebox{90}{Crash}  & \rotatebox{90}{Frame}
              & &
              & \rotatebox{90}{Fog}    & \rotatebox{90}{Wet}
              & \rotatebox{90}{Snow}   & \rotatebox{90}{Motion}
              & \rotatebox{90}{Beam}   & \rotatebox{90}{Cross}
              & \rotatebox{90}{Echo}   & \rotatebox{90}{Sensor}
              & \\
        \midrule
        \rowcolor{gray!10}\multicolumn{23}{@{}l}{\textbf{OccStress-nuScenes}} \\
        $\mathrm{C}$ & SparseOcc~\cite{tang2024sparseocc} & 21.7 & 14.8 &  9.2 & 12.5 &  3.2 &  7.8 & 11.8 & 10.0 &  8.4 &  9.7 & 10.91 & \multicolumn{10}{c}{--} \\
        $\mathrm{C}$ & FB-OCC~\cite{li2023fb}             & 39.3 & 28.0 & 16.9 & 27.6 &  8.0 & 23.7 & 25.2 & 23.6 & 25.4 & 22.3 & 24.41 & \multicolumn{10}{c}{--} \\
        $\mathrm{C}$ & ViewFormer~\cite{li2024viewformer} & \underline{41.8} & 28.0 & 19.5 & 29.9 &  6.3 & 23.8 & 25.9 & 23.3 & 25.3 & 22.8 & 25.08 & \multicolumn{10}{c}{--} \\
        $\mathrm{C}$ & COTR~\cite{ma2024cotr}             & 39.2 & 31.4 & 19.6 & 29.8 &  9.3 & 27.9 & 28.1 & 18.8 & 18.7 & 22.9 & 24.98 & \multicolumn{10}{c}{--} \\
        $\mathrm{C}$ & PanoOcc~\cite{wang2024panoocc}     & 36.6 & \underline{34.2} & \underline{25.5} & 31.8 & \underline{13.7} & 24.6 & \underline{30.3} & 21.2 & 18.8 & 25.0 & 26.69 & \multicolumn{10}{c}{--} \\
        $\mathrm{C}$ & ALOcc~\cite{chen2025alocc}         & \textbf{44.0} & \textbf{35.0} & 24.2 & \textbf{35.1} &  9.6 & \textbf{30.5} & 30.2 & \underline{27.7} & \underline{29.1} & \underline{27.7} & \underline{29.92} & \multicolumn{10}{c}{--} \\
        $\mathrm{C}$ & STCOcc~\cite{liao2025stcocc}       & 41.7 & 34.0 & \textbf{30.9} & \underline{33.3} & \textbf{21.1} & \underline{30.2} & \textbf{30.7} & \textbf{31.2} & \textbf{30.3} & \textbf{30.2} & \textbf{31.96} & \multicolumn{10}{c}{--} \\
        \addlinespace[1pt]
        $\mathrm{C{+}L}$ & OccFusion~\cite{zhang2024occfusion} & 23.8 & 23.0 & 13.6 & 22.4 & 11.9 & 17.2 & 19.1 & 15.6 & 14.2 & 17.1 & 18.12 & 16.5 & 22.9 & 21.1 & 21.2 & 18.8 & 19.9 & 20.8 & 14.3 & 19.4 & 20.14 \\
        $\mathrm{C{+}L}$ & FusionOcc~\cite{zhang2024fusionocc} & \underline{45.4} & \underline{40.6} & \underline{23.5} & \underline{42.2} & \textbf{28.1} & \underline{34.0} & \underline{29.0} & \underline{25.7} & \textbf{26.2} & \underline{31.2} & \underline{33.28} & \textbf{37.1} & \underline{45.7} & \underline{44.3} & \underline{36.8} & \textbf{44.4} & \underline{41.5} & \underline{44.7} & \textbf{43.1} & \underline{42.2} & \underline{42.80} \\
        $\mathrm{C{+}L}$ & SDGOcc~\cite{duan2025sdgocc} & \textbf{51.6} & \textbf{45.6} & \textbf{33.0} & \textbf{45.0} & \underline{24.7} & \textbf{42.2} & \textbf{42.5} & \textbf{26.2} & \underline{24.3} & \textbf{35.4} & \textbf{37.77} & \underline{31.9} & \textbf{50.2} & \textbf{47.7} & \textbf{39.3} & \underline{43.3} & \textbf{47.1} & \textbf{46.2} & \underline{35.3} & \textbf{42.6} & \textbf{44.14} \\
        \midrule
        \rowcolor{gray!10}\multicolumn{23}{@{}l}{\textbf{OccStress-Waymo}} \\
        $\mathrm{C}$ & CVT-Occ~\cite{ye2024cvtocc}
            & 25.6
            & 20.9 & 14.4 & 18.2 & 5.8 & 16.0 & 18.0 & 9.9 & 12.6
            & 14.5 & 15.86
            & \multicolumn{10}{c}{--} \\
        $\mathrm{C{+}L}$ & EFFOcc~\cite{effocc2025shi}
            & 49.6
            & 48.8 & 48.4 & 47.8 & 48.1 & 48.9 & 48.7 & 32.6 & 34.4
            & 44.7 & 45.61
            & 29.8 & 49.0 & 37.7 & 30.6 & 43.3 & 35.4 & 47.3 & 43.4
            & 39.6 & 41.22 \\
        \midrule
        \rowcolor{gray!10}\multicolumn{23}{@{}l}{\textbf{OccStress-CARLA}} \\
        $\mathrm{C}$ & FlashOcc~\cite{yu2023flashocc}
            & 18.7
            & 16.4 & 11.9 & 15.4 & 13.0 & 14.0 & 15.4 & 5.6 & 8.4
            & 12.5 & 13.39
            & \multicolumn{10}{c}{--} \\
        $\mathrm{C{+}L}$ & EFFOcc~\cite{effocc2025shi}
            & 16.2
            & 13.2 & 11.3 & 15.4 & 11.3 & 14.4 & 15.5 & 12.1 & 13.3
            & 13.3 & 13.81
            & 15.5 & 16.2 & 15.9 & 15.7 & 15.8 & 15.4 & 16.2 & 15.3
            & 15.7 & 15.84 \\
        \bottomrule
    \end{tabular}
    \end{adjustbox}
\end{table*}

\subsection{Experimental Setup}

Experiments use OccStress-nuScenes, OccStress-Waymo, and OccStress-CARLA, with 4,519, 5,978, and 330 strict H4/F6 anchors, respectively. Each 4D sample contains four historical occupancy states, the current state, and six future targets sampled every 0.5 seconds. The primary 4D metric is semantic mIoU averaged over the 1s, 2s, and 3s forecasting horizons. Clean and corrupted protocols use the same anchor samples. For OccStress-Waymo and OccStress-CARLA, the forecasting checkpoints are evaluated zero-shot, while the upstream 3D occupancy sources are dataset-native.
Model input windows and cross-dataset adapters are specified in Appendix~\ref{sec:app_transfer_setup}.
We use the aggregate metrics defined in Sec.~\ref{sec:evaluation_metrics}: \(mRS\) is the primary robustness score, \(CRF_2\) balances clean and corrupted-condition performance on the mIoU scale, and \(UCR\) measures upstream clean-state quality. Experiments use NVIDIA RTX 5090 and A100 GPUs.

\subsection{Robustness of Upstream 3D Occupancy Sources}

Table~\ref{tab:2_3d_occ_robustness} evaluates candidate 3D occupancy sources
under observation corruptions before they are used in the downstream 4D
pipeline. The \(\mathrm{C}\) and \(\mathrm{C{+}L}\) rows define different input
settings and are therefore compared separately within each dataset. For a
camera--LiDAR source, an image corruption perturbs only the image stream while
leaving point-cloud evidence available, and a point-cloud corruption follows
the converse setting.

The source-selection criteria are \(RS^{I}_{3D}\) for image corruptions and
\(RS^{P}_{3D}\) for point-cloud corruptions, each computed within the
corresponding input-modality block. For \(S\in\{I,P\}\), the source-local
balanced score is
\(CRF_2^S=5CS\,RS^S_{3D}/(4CS+RS^S_{3D})\). It is reported as a diagnostic,
while source selection remains based on absolute corrupted occupancy quality.
STCOcc is selected as the
image-corruption source because it has the highest \(RS^{I}_{3D}\) among
camera-only models, whereas SDGOcc is selected as the point-cloud-corruption
source because it has the highest \(RS^{P}_{3D}\) among camera--LiDAR models.
Their clean and corrupted occupancy predictions define the upstream inputs for
4D forecasting.

For OccStress-Waymo, CVT-Occ~\cite{ye2024cvtocc} and EFFOcc~\cite{effocc2025shi} provide the dataset-native
camera-only and camera--LiDAR sources, respectively. CVT-Occ is evaluated under
image corruptions, while EFFOcc is evaluated separately under image and
point-cloud corruptions with the other modality kept clean. Every clean and
corrupted source setting covers all 202 scenes and 7,998 evaluation frames.

\definecolor{avgshade}{gray}{0.93}

\Needspace{14\baselineskip}
\begin{wraptable}[13]{r}{0.5\textwidth}
    \vspace{-1.2em}
    \centering
    \scriptsize
    \setlength{\tabcolsep}{1.3pt}
    \renewcommand{\arraystretch}{1.08}
    \captionsetup{font=small,justification=raggedright,singlelinecheck=false}
    \caption{\textbf{4D robustness on OccStress-nuScenes}. %
    Scores are
    mIoU (\%) averaged over 1s, 2s, and 3s. Non-traffic corruption scores average
    three severities and three protocols. Traffic uses full-sequence mirroring.}
    \label{tab:3_4d_occ_manual}
    \begin{adjustbox}{max width=\linewidth}
    \begin{tabular}{@{}l c ccccc@{}}
        \toprule
        \multirow{2}{*}{\textbf{Method}}
            & \multirow{2}{*}{$\textbf{CS}^{\textit{O}}$}
            & \multicolumn{5}{c}{\textbf{4D Occupancy Corruptions}} \\
        \cmidrule(lr){3-7}
            & & Semantic & Hole & Dropout & Misaligned & Traffic \\
        \midrule
        OccWorld~\cite{zheng2024occworld}
            & 17.21
            & 14.6 & 17.0 & 14.1 & 17.2 & 16.3
            \\
        DOME~\cite{gu2024dome}
            & 26.75
            & 22.5 & 26.6 & 22.8 & 14.1 & 17.5
            \\
        COME~\cite{shicome}
            & 34.52
            & 29.0 & 34.3 & 29.7 & 34.5 & 19.1
            \\
        I$^{2}$-World~\cite{liao2025i2}
            & \underline{39.78}
            & \underline{32.9} & \underline{39.3} & \underline{34.3} & \underline{38.4} & \textbf{38.1}
            \\
        GenieDrive~\cite{yang2026geniedrive}
            & \textbf{42.60}
            & \textbf{35.4} & \textbf{42.0} & \textbf{37.1} & \textbf{42.4} & \underline{22.7}
            \\
        \bottomrule
    \end{tabular}
    \end{adjustbox}
\end{wraptable}

For OccStress-CARLA, FlashOcc-CARLA and EFFOcc-CARLA provide the dataset-native
camera-only and camera--LiDAR sources, respectively. Both are adapted and
trained on the released CARLA training split rather than reported as
paper-reproduction numbers. FlashOcc-CARLA is evaluated under image
corruptions, while EFFOcc-CARLA is evaluated under both image and point-cloud
corruptions with the other modality kept clean. Every clean and corrupted
source setting covers all 360 validation frames. Because the released
validation labels contain only five occupied classes, CARLA source-level
semantic mIoU averages present classes.

\subsection{Robustness of Manual Occupancy-State Corruptions}

Table~\ref{tab:3_4d_occ_manual} evaluates controlled corruptions applied
directly to the occupancy-state sequence. For non-traffic corruptions, scores
average over three severities and the \textit{Current-only}, \textit{Recent-burst}, and
\textit{History-only} protocols.
Semantic and dropout corruptions produce the largest non-traffic degradations
for OccWorld, COME, I$^{2}$-World, and GenieDrive, whereas DOME is uniquely
sensitive to misalignment. Hole is consistently mild. These results establish the
manual state-corruption track as a controlled counterpart to the upstream
pipeline track.
An external-validity audit on OccStress-nuScenes compares these interventions with
occupancy residuals from clean STCOcc and SDGOcc predictions. On I$^2$-World, the matched
manual and residual-induced six-horizon response shapes have cosine similarity
of 0.956--1.000, while the residual effects are only 17--51\% as large,
supporting the manual track as a diagnostic stress envelope rather than a
calibrated deployment distribution (Appendix~\ref{sec:app_manual_external_validity}).
Traffic is treated separately as a full-sequence transformation rather than a
destructive perturbation. 
OccWorld and I$^{2}$-World remain close to
clean, while DOME, COME, and GenieDrive degrade under the mirrored layout. The
largest drops occur for COME and GenieDrive.

\subsection{4D Forecasting Robustness under OccStress}

\definecolor{avgshade}{gray}{0.93}
\newcommand{\rsdeltaGain}[1]{\textcolor{red}{\scriptsize(+#1)}}
\newcommand{\rsdeltaDrop}[1]{\textcolor[RGB]{96,177,87}{\scriptsize(-#1)}}

\begin{table*}[t]
    \centering
    \small
    \setlength{\tabcolsep}{3.5pt}
    \renewcommand{\arraystretch}{1.20}
    \caption{\textbf{Robustness summary of 4D occupancy forecasting in
    OccStress}. Semantic mIoU scores (\%) average the 1s, 2s, and 3s
    forecasting horizons. \(UCR\) is dimensionless.
    \(CS\) denotes clean score. \(I/P/O\) denotes
    image, point-cloud, and occupancy-state corruption sources.
    \textit{Cur/Burst/Hist} denotes the three temporal protocols.
    \(\forall\) averages the dimensions not named by a column. Parentheses in
    \(mRS\) report \(mRS{-}CS^{O}\). Bold and underlined denote the
    best and second-best results within each dataset block.}
    \label{tab:4_occstress_summary}
    \begin{adjustbox}{max width=\textwidth}
    \begin{tabular}{@{}l ccc ccc ccc cc>{\columncolor{avgshade}}c@{}}
        \toprule
        \multirow{2}{*}{\textbf{Method}} 
        & \multicolumn{3}{c}{\textbf{Clean}}
        & \multicolumn{3}{c}{\textbf{Source-Specific}}
        & \multicolumn{3}{c}{\textbf{Protocol-Specific}}
        & \multicolumn{3}{c}{\textbf{Summary}} \\
        \cmidrule(lr){2-4} \cmidrule(lr){5-7} \cmidrule(lr){8-10} \cmidrule(lr){11-13}
        & $\textbf{CS}^{\textit{I}}$
        & $\textbf{CS}^{\textit{P}}$
        & $\textbf{CS}^{\textit{O}}$
        & $\textbf{RS}^{\textit{I}}_{\forall}$
        & $\textbf{RS}^{\textit{P}}_{\forall}$
        & $\textbf{RS}^{\textit{O}}_{\forall}$
        & $\textbf{RS}^{\forall}_{\textit{Cur}}$
        & $\textbf{RS}^{\forall}_{\textit{Burst}}$
        & $\textbf{RS}^{\forall}_{\textit{Hist}}$
        & $\textbf{UCR}$
        & $\textbf{mRS}$
        & $\textbf{CRF}_{\mathbf{2}}$ \\
        \midrule
        \multicolumn{13}{@{}l}{\textbf{OccStress-nuScenes}} \\
        \addlinespace[1pt]
        OccWorld~\cite{zheng2024occworld}
            & 10.19
            & 11.03 & 17.21 & 8.54 & 9.81 & 15.71
            & 11.07 & 11.13 & 11.85
            & \textbf{0.62} & 11.35\,\rsdeltaDrop{5.86} & 12.18
            \\
        DOME~\cite{gu2024dome}
            & 13.68
            & 15.73 & 26.75 & 12.24 & 14.37 & 21.53
            & 15.55 & 15.11 & 17.47
            & \underline{0.55} & 16.05\,\rsdeltaDrop{10.70} & 17.44
            \\
        COME~\cite{shicome}
            & 15.58
            & 19.56 & 34.52 & 13.75 & 17.93 & 31.87
            & 20.70 & 20.40 & 22.45
            & 0.51 & 21.18\,\rsdeltaDrop{13.33} & 22.96
            \\
        I$^{2}$-World~\cite{liao2025i2}
            & \underline{17.28}
            & \underline{21.44} & \underline{39.78} & \underline{15.31} & \underline{19.33} & \underline{36.24}
            & \underline{22.46} & \underline{22.46} & \underline{25.96}
            & 0.49 & \underline{23.63}\,\rsdeltaDrop{16.15} & \underline{25.71}
            \\
        GenieDrive~\cite{yang2026geniedrive}
            & \textbf{17.79}
            & \textbf{23.37} & \textbf{42.60} & \textbf{15.81} & \textbf{21.18} & \textbf{39.23}
            & \textbf{24.32} & \textbf{24.24} & \textbf{27.66}
            & 0.48 & \textbf{25.40}\,\rsdeltaDrop{17.20} & \textbf{27.64}
            \\
        \midrule
        \multicolumn{13}{@{}l}{\textbf{OccStress-Waymo}~\textit{(zero-shot forecasters)}} \\
        \addlinespace[1pt]
        OccWorld~\cite{zheng2024occworld}
            & 3.92
            & 6.18 & 18.36 & 2.85 & 5.31 & 17.30
            & 8.30 & 8.35 & 8.81
            & \textbf{0.28} & 8.49\,\rsdeltaDrop{9.88} & 9.51
            \\
        DOME~\cite{gu2024dome}
            & 2.55
            & 3.38 & 24.68 & 2.12 & 3.24 & 20.87
            & 8.50 & 8.29 & 9.45
            & 0.12 & 8.75\,\rsdeltaDrop{15.94} & 10.04
            \\
        COME~\cite{shicome}
            & 2.49
            & 3.79 & 35.01 & 2.13 & 3.63 & 30.10
            & 11.60 & 11.29 & 12.97
            & 0.09 & 11.95\,\rsdeltaDrop{23.06} & 13.77
            \\
        I$^{2}$-World~\cite{liao2025i2}
            & \underline{4.96}
            & \underline{9.74} & \underline{35.87} & \underline{3.75} & \underline{8.60} & \underline{33.85}
            & \underline{14.73} & \underline{14.71} & \underline{16.76}
            & 0.20 & \underline{15.40}\,\rsdeltaDrop{20.47} & \underline{17.39}
            \\
        GenieDrive~\cite{yang2026geniedrive}
            & \textbf{6.49}
            & \textbf{10.72} & \textbf{39.95} & \textbf{4.95} & \textbf{9.65} & \textbf{38.03}
            & \textbf{16.94} & \textbf{16.84} & \textbf{18.86}
            & \underline{0.22} & \textbf{17.54}\,\rsdeltaDrop{22.41} & \textbf{19.76}
            \\
        \midrule
        \multicolumn{13}{@{}l}{\textbf{OccStress-CARLA}~\textit{(zero-shot forecasters)}} \\
        \addlinespace[1pt]
        OccWorld~\cite{zheng2024occworld}
            & 4.78
            & 6.79 & 23.46 & 4.55 & 6.00 & 21.90
            & 10.87 & 10.81 & 10.77
            & \underline{0.25} & 10.82\,\rsdeltaDrop{12.64} & 12.12
            \\
        DOME~\cite{gu2024dome}
            & 7.30
            & 8.60 & 30.26 & 5.75 & 8.46 & 26.17
            & 13.22 & 12.79 & 14.36
            & \textbf{0.26} & 13.46\,\rsdeltaDrop{16.81} & 15.14
            \\
        COME~\cite{shicome}
            & \underline{8.20}
            & \underline{9.35} & \underline{44.37} & \underline{6.19} & \underline{9.19} & \underline{39.61}
            & \underline{17.78} & \underline{17.36} & \underline{19.84}
            & 0.20 & \underline{18.33}\,\rsdeltaDrop{26.05} & \underline{20.77}
            \\
        I$^{2}$-World~\cite{liao2025i2}
            & 7.02
            & 9.14 & 34.38 & 5.91 & 8.88 & 33.52
            & 15.80 & 15.78 & 16.73
            & 0.24 & 16.10\,\rsdeltaDrop{18.28} & 18.02
            \\
        GenieDrive~\cite{yang2026geniedrive}
            & \textbf{9.93}
            & \textbf{9.84} & \textbf{53.34} & \textbf{7.20} & \textbf{9.59} & \textbf{51.35}
            & \textbf{22.05} & \textbf{21.90} & \textbf{24.18}
            & 0.19 & \textbf{22.71}\,\rsdeltaDrop{30.63} & \textbf{25.66}
            \\
        \bottomrule
    \end{tabular}
    \end{adjustbox}
\end{table*}

Table~\ref{tab:4_occstress_summary} is the aggregate view of OccStress. It
combines source-specific robustness over image, point-cloud, and
occupancy-state corruption sources with protocol-specific robustness over the
\textit{Current-only}, \textit{Recent-burst}, and
\textit{History-only} temporal settings. We regard
\(mRS\) as the primary aggregate robustness score. Since retention ratios can
favor models with low clean baselines, we report \(CRF_2\) as a robust-biased
F-score between \(CS^O\) and \(mRS\). \(CRF_2\) preserves the mIoU scale while
requiring both strong clean forecasting and strong corrupted-condition
performance.

Under this view, GenieDrive achieves the highest \(mRS\) and \(CRF_2\) on all
three datasets. I$^{2}$-World ranks second on OccStress-nuScenes and OccStress-Waymo, while COME
ranks second on OccStress-CARLA. Unlike retention ratios, \(CRF_2\) does not reward a
large fraction of a weaker clean baseline.
The \(UCR\) values
further show that clean upstream occupancy remains a major bottleneck.

  \section{Diagnostic Analysis}

\subsection{Error Source Matters}

Tables~\ref{tab:2_3d_occ_robustness}, \ref{tab:3_4d_occ_manual}, and \ref{tab:4_occstress_summary} separate robustness by where errors enter the occupancy forecasting chain.
On OccStress-nuScenes, the best clean camera-only estimator is not the most robust under image corruptions.

After the upstream predictions are passed to 4D forecasters, Table~\ref{tab:4_occstress_summary} shows that point-cloud-induced upstream states consistently outperform image-induced upstream states across all three datasets and five forecasters.
Both upstream sources nevertheless remain below the manual occupancy-state source in every dataset--model block, consistent with upstream state quality being a major robustness bottleneck.
Separately, for I$^2$-World on OccStress-nuScenes, a 10,000-sample scene-clustered paired bootstrap gives
\(CS^O-mRS=16.15\) mIoU with a 95\% confidence interval of
\([15.50,16.81]\), confirming an overall clean-to-stressed gap under
validation-scene resampling (Appendix~\ref{sec:app_uncertainty}).

Manual corruptions remove the perception model from the evaluation loop and directly test how 4D forecasters respond to erroneous temporal states. Table~\ref{tab:3_4d_occ_manual} shows that semantic replacement and dropout are the most damaging non-traffic corruptions for four forecasters, while DOME is particularly sensitive to misalignment. Traffic mirroring instead probes structured layout transfer. OccWorld and I\(^2\)-World remain close to their clean references, whereas DOME, COME, and GenieDrive degrade under the mirrored layout.

\subsection{History Can Stabilize or Propagate Errors}

Table~\ref{tab:4_occstress_summary} shows that History-only is the least damaging aggregate protocol in 14 of 15 dataset--model blocks under these predefined failure regimes.
The manual results show the same ordering for COME, with Cur/Burst/Hist scores of 31.15/30.80/33.67 mIoU.
Reporting the protocols separately prevents these differences from being hidden by a single average.

\subsection{Temporal Error Propagation across Horizons}

\begin{wrapfigure}[15]{r}{0.3\textwidth}
  \vspace{-1.6em}
  \centering
  \includegraphics[width=\linewidth]{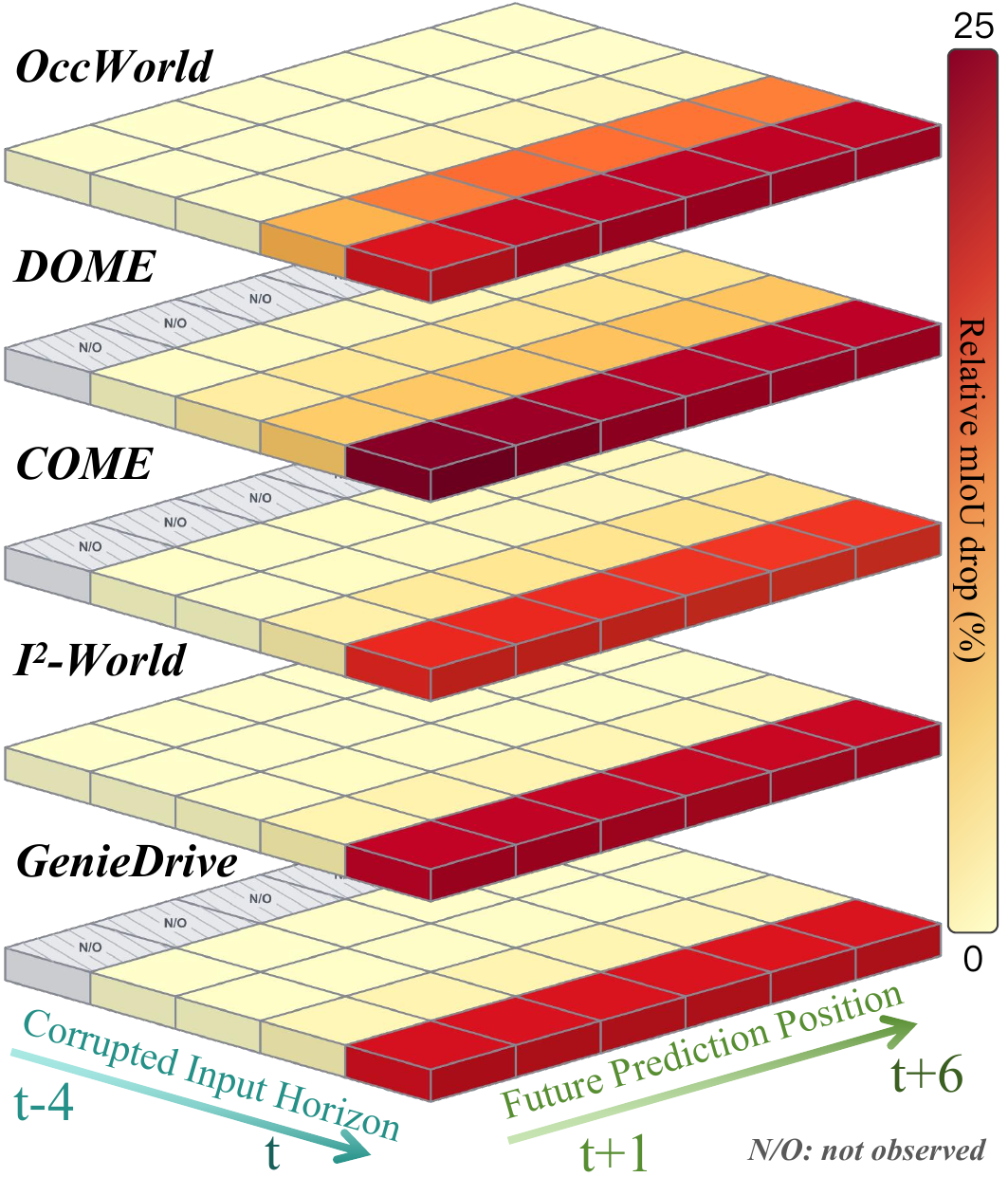}
  \captionsetup{font=small,justification=raggedright,singlelinecheck=false}
  \caption{Position-wise temporal sensitivity.}
  \label{fig:position_sweep}
  \vspace{-0.2em}
\end{wrapfigure}

To inspect temporal sensitivity beyond the three aggregate protocols, we
corrupt one input state at a time while keeping all other inputs and future
targets clean. Figure~\ref{fig:position_sweep} reports relative semantic mIoU
drops averaged over ten shared hard-corruption settings and the 4,519-anchor
OccStress-nuScenes evaluation set. Five input positions are shown, with N/O marking
unobserved inputs. Cross-model statistics use the four shared positions.
Position-wise means average relative drops computed separately at the six
future horizons (0.5--3.0 seconds).
Across these methods, corrupting the current state is
most damaging, with mean relative drops of 17.46--22.74\%, compared with
0.08--1.19\% at the earliest common historical position. OccWorld and DOME retain greater
sensitivity to recent history, whereas I$^2$-World and GenieDrive are more
sharply anchored to the current state.
Separately, across three pre-specified I$^2$-World position sweeps, degradation
at the current position exceeds the mean historical-position degradation by
\(7.08\) mIoU, with a 95\% confidence interval of \([6.74,7.42]\).

\subsection{Qualitative Cases}

\begin{figure}[t]
    \centering
    \includegraphics[width=1.0\linewidth]{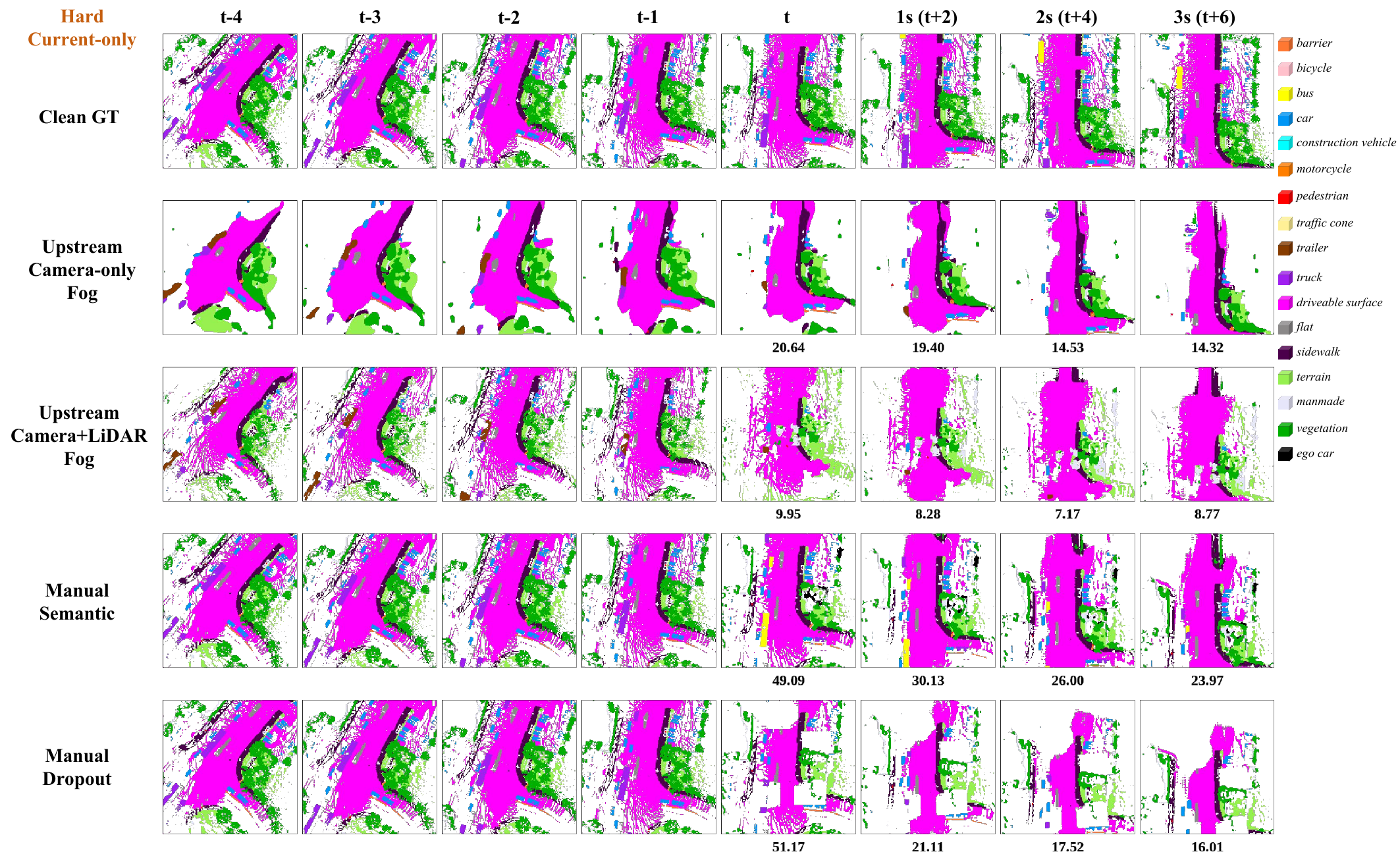} 
    \caption{
    Qualitative cases of I$^2$-World on OccStress-nuScenes under the hard \textit{Current-only} protocol.
    }
    \label{fig:vis}
    \vspace{-0.2em}
  \end{figure}

Figure~\ref{fig:vis} shows representative I$^2$-World failures under upstream and manual corruptions, both of which produce prediction errors that persist across future horizons.

\subsection{Case Study: History Fusion Mechanism Matters}

\begin{wrapfigure}[4]{r}{0.4\textwidth}
  \vspace{-3.2em}
  \centering
  \includegraphics[width=\linewidth]{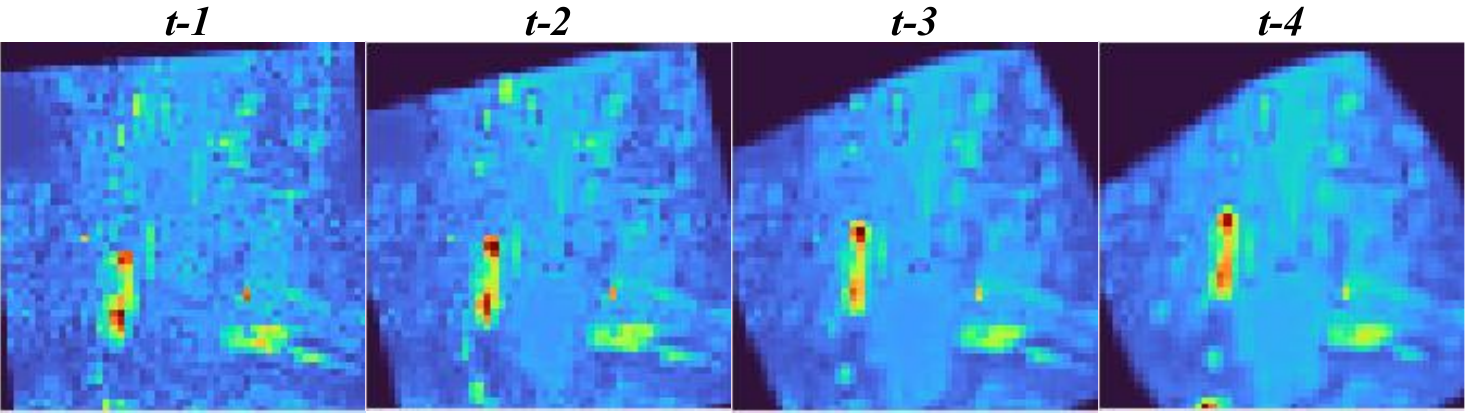}
  \captionsetup{font=small,justification=raggedright,singlelinecheck=false}
  \caption{I$^2$-World slot latents.}
  \label{fig:slot_latent}
  \vspace{-0.2em}
\end{wrapfigure}

We inspect I$^2$-World on OccStress-nuScenes to determine what its history slots contribute before forecasting (Figure~\ref{fig:slot_latent}).
The original inter branch fuses ego-aligned historical BEV slots with the current residual, but the alignment operation itself introduces a bilinear-like smoothing effect. 
To isolate this factor, we replace real historical slots with low-pass filtered copies of the current latent and vary the bilinear strength while keeping the stage-1 reconstruction objective unchanged. 
The $mIoU$ scores for \textit{origin}, strength $(0,0,0,0)$, $(0,0.25,0.5,0.75)$, and $(0,0,1,2)$ are $81.88$, $83.42$~\textcolor{red}{\scriptsize(+1.54)}, $83.56$~\textcolor{red}{\scriptsize(+1.68)}, and $83.82$~\textcolor{red}{\scriptsize(+1.94)}, respectively.
Although these slots contain no real temporal history, they improve stage-1 reconstruction. This suggests that the gain partly arises from reconstruction-friendly smoothing and capacity rather than explicit historical information.

The evaluated methods also use different representation bottlenecks. OccWorld adopts a frame-wise discrete VQ-VAE, I$^2$-World applies discrete tokenization after temporal alignment and aggregation, DOME and COME share a continuous KL-regularized Occ-VAE, and GenieDrive uses a frame-wise continuous tri-plane VAE. Discrete tokenization can introduce abrupt changes near codebook boundaries, whereas continuous latents avoid hard quantization but may still suppress fine geometric details through compression and KL regularization. Temporal encoding can average independent noise, but it can also spread localized corruption across latent states or reconstructed frames. We therefore interpret the observed robustness as an end-to-end property rather than causal evidence for a particular VAE type without controlled VAE ablations.

  \section{Conclusion}

We present \textbf{OccStress}, a stress-testing benchmark for the 4D occupancy forecasting chain. Instead of treating robustness as a frame-level perception problem, OccStress evaluates how errors enter through upstream observations or occupancy states, persist in temporal context, and affect future forecasts. Through \textit{upstream} and \textit{manual} tracks, three temporal protocols, and source-wise and protocol-wise robustness scores, OccStress exposes failure modes that are hidden by clean forecasting evaluation. Across OccStress-nuScenes, OccStress-Waymo, and OccStress-CARLA with five forecasters, our experiments show that clean accuracy alone does not reveal source- and position-specific sensitivity, upstream occupancy quality is a major bottleneck, and current forecasters exhibit different temporal sensitivity patterns. We hope OccStress provides a reusable evaluation asset for developing more reliable occupancy forecasting models and more robust temporal fusion mechanisms.

  \clearpage
  \noindent \textbf{Acknowledgment} This work was supported in part by the National Natural Science Foundation of China under Grant No. 62503166, in part by the Hunan Provincial Research and Development Project under Grant number 2026QK3018, and in part by the Yuelushan Industrial Innovation Center. 
  { \small \bibliographystyle{plain} \bibliography{main} }

  \clearpage
\appendix
\raggedbottom

\newcommand{\curprot}{Cur}
\newcommand{\burstprot}{Burst}
\newcommand{\histprot}{Hist}
\definecolor{occOthers}{RGB}{0,0,0}
\definecolor{occBarrier}{RGB}{255,120,50}
\definecolor{occBicycle}{RGB}{255,192,203}
\definecolor{occBus}{RGB}{255,255,0}
\definecolor{occCar}{RGB}{0,150,245}
\definecolor{occConstruction}{RGB}{0,255,255}
\definecolor{occMotorcycle}{RGB}{255,127,0}
\definecolor{occPedestrian}{RGB}{255,0,0}
\definecolor{occTrafficCone}{RGB}{255,240,150}
\definecolor{occTrailer}{RGB}{135,60,0}
\definecolor{occTruck}{RGB}{160,32,240}
\definecolor{occDriveable}{RGB}{255,0,255}
\definecolor{occOtherFlat}{RGB}{139,137,137}
\definecolor{occSidewalk}{RGB}{75,0,75}
\definecolor{occTerrain}{RGB}{150,240,80}
\definecolor{occManmade}{RGB}{230,230,250}
\definecolor{occVegetation}{RGB}{0,175,0}
\newcommand{\occbox}[1]{\raisebox{0.15ex}{\textcolor{#1}{\rule{0.85ex}{0.85ex}}}}
\newcommand{\occhead}[2]{\rotatebox[origin=c]{90}{\scriptsize\occbox{#1}\hspace{0.25em}#2\strut}}

\section{Limitations}

OccStress is designed as a controlled robustness benchmark for the 4D
occupancy forecasting chain, and its scope is intentionally bounded. First, the
current release spans OccStress-nuScenes, OccStress-Waymo, and OccStress-CARLA, but it does
not fully cover geographic diversity, rare traffic rules, or long-tail driving
events; CARLA is limited to three validation scenes and five occupied classes,
and Waymo/CARLA forecasting is evaluated zero-shot. Second, the corruption families are selected to
represent common sensor, upstream-perception, and occupancy-state failure
modes, but they do not exhaust all deployment failures such as hardware faults,
map errors, calibration drift over long periods, or closed-loop planning
feedback. Third, the upstream 4D evaluation uses representative 3D occupancy
sources selected from the completed robust-source study; adding more camera,
LiDAR, and multi-modal occupancy estimators may change the absolute source
quality available to downstream forecasters.

\section{Broader Impacts}

OccStress is intended to support robustness-oriented evaluation of 3D and 4D
occupancy systems before they are considered in safety-critical autonomy
pipelines. By exposing how perception corruptions and occupancy-state errors
propagate through forecasting models, the benchmark can help researchers
identify brittle failure modes, compare methods under shared stress conditions,
and prioritize improvements that are not visible from clean-set accuracy alone.
The benchmark is evaluation-only: it does not provide a safety certificate, does
not replace closed-loop testing, and should not be used as evidence that a model
is ready for deployment. A potential negative impact is benchmark overfitting or
overconfidence from reporting high scores on a fixed set of corruptions. We
therefore present OccStress as a diagnostic complement to broader validation
covering additional geographies, sensors, traffic conventions, rare events, and
closed-loop interactions.

\section{Additional OccStress Evaluation Details}

This appendix provides the detailed tables used to audit the compact results in
the main paper. Unless otherwise specified, detailed breakdowns use OccStress-nuScenes, while the cross-dataset aggregate is reported in Table~\ref{tab:4_occstress_summary}. Unless otherwise stated, occupancy scores are semantic
mIoU (\%). For 4D occupancy forecasting, scores are averaged over the 1s, 2s,
and 3s forecasting horizons. Tables explicitly marked as IoU report
horizon-averaged binary occupancy IoU under the same protocol aggregation. The
protocol names are abbreviated as
\curprot{} for \textit{Current-only}, \burstprot{} for \textit{Recent-burst}, and
\histprot{} for \textit{History-only}. In metric tables, bold and underlined
entries mark the best and second-best scores within the corresponding
comparable column group.

\subsection{Cross-Dataset and Temporal Setup}
\label{sec:app_transfer_setup}

All forecasting checkpoints are trained on nuScenes and transferred to Waymo
and CARLA without target-dataset fine-tuning. The upstream estimators are
dataset-native, including the CARLA adaptations trained on its released training
split. The common H4/F6 inventory requires valid states at
\(t-2.0,t-1.5,t-1.0,t-0.5,t\) seconds and six future targets through \(t+3.0\) seconds.
OccWorld and I$^2$-World use all five input states. COME, DOME, and GenieDrive
retain their four-state conditioning window from \(t-1.5\) seconds to \(t\).
Their unused earliest state is excluded from cross-model position averages.

Waymo and CARLA occupancy labels are mapped into the checkpoints' Occ3D
18-label vocabulary, including free space, on a \(200\times200\times16\) grid.
The Waymo map is
\(\{0{:}0,1{:}4,2{:}7,3{:}15,4{:}2,5{:}15,6{:}15,7{:}8,
8{:}2,9{:}6,10{:}15,11{:}16,12{:}16,13{:}11,14{:}13,23{:}17\}\).
The CARLA map is
\(\{0{:}0,1{:}4,2{:}2,3{:}6,4{:}7,5{:}8,6{:}16,7{:}11,8{:}14,9{:}15,10{:}17\}\).
CARLA occupancy and camera-visibility masks are reflected along local \(y\),
with the corresponding transformation applied to poses and controls, to obtain
the right-handed model input convention. Cross-dataset forecasting mIoU
averages occupied classes present in the evaluation targets and excludes free
space. CARLA validation contains car, driveable surface, terrain, manmade, and
vegetation. Classes absent from the released targets are not synthesized.

\subsection{Benchmark Generation}

OccStress is generated by first defining a clean temporal backbone and then
instantiating corrupted evaluation records on top of the same anchors. Let
\(\mathbf{O}_{t}\) denote the occupancy state at time \(t\). For each anchor
\(t\), the clean clip contains \(H\) historical states, the current state, and
\(F\) future targets:
\begin{equation}
    \mathbf{X}_{t}
    =
    \{\mathbf{O}_{t-H},\ldots,\mathbf{O}_{t-1},\mathbf{O}_{t}\},
    \qquad
    \mathbf{Y}_{t}
    =
    \{\mathbf{O}_{t+1},\ldots,\mathbf{O}_{t+F}\}.
\end{equation}
Only anchors with a complete input and target window are retained:
\begin{equation}
    \mathcal{A}
    =
    \{t \mid \mathbf{O}_{t-H:t}\ \text{and}\ \mathbf{O}_{t+1:t+F}
    \ \text{are valid}\}.
\end{equation}

For a corruption setting \(q=(S,c,\ell,p)\), \(S\) specifies the corruption
source, \(c\) the corruption family, \(\ell\) the severity level, and \(p\) the
temporal protocol. The corruption operator \(\Phi_{S,c,\ell}\) is interpreted
as either the composed upstream perception error or a direct occupancy-state
perturbation, depending on the source. Its temporal placement is controlled by
a binary mask over the input window, where \(i=0\) denotes the current state and
\(i>0\) denotes the \(i\)-th previous history state, and \(K\) counts the
history states in the recent burst. We set \(K=1\), corrupting the current
state and its nearest history state:
\begin{equation}
    M_{p}(i)=
    \begin{cases}
    1, & p=\mathrm{Cur}\ \text{and}\ i=0,\\
    1, & p=\mathrm{Burst}\ \text{and}\ 0\leq i\leq K,\\
    1, & p=\mathrm{Hist}\ \text{and}\ 1\leq i\leq H,\\
    0, & \text{otherwise}.
    \end{cases}
\end{equation}
The resulting corrupted temporal input is
\begin{equation}
    \tilde{\mathbf{O}}^{q}_{t-i}
    =
    \begin{cases}
    \Phi_{S,c,\ell}(\mathbf{O}_{t-i}), & M_{p}(i)=1,\\
    \mathbf{O}^{S,\mathrm{clean}}_{t-i}, & M_{p}(i)=0.
    \end{cases}
\end{equation}
Here \(\mathbf{O}^{S,\mathrm{clean}}\) is a clean-input prediction from the same
upstream estimator for \(S\in\{I,P\}\), and ground truth for \(S=O\).
The 21 families comprise eight image, eight point-cloud, and five manual
families. Three severities for each of the 20 non-traffic families plus one
traffic transformation give 61 configurations, before temporal protocols.
For ordinary corruptions, future targets remain clean so that performance drops
are attributable to corrupted inputs. Traffic mirroring is the exception: it
applies a shared left-right transformation \(T_{\mathrm{traffic}}\) to inputs
and targets to preserve a consistent traffic layout:
\begin{equation}
    \tilde{\mathbf{Y}}^{q}_{t}
    =
    \begin{cases}
    T_{\mathrm{traffic}}(\mathbf{Y}_{t}), & c=\mathrm{traffic},\\
    \mathbf{Y}_{t}, & \text{otherwise}.
    \end{cases}
\end{equation}
Each benchmark item is therefore a source-conditioned protocol record
\begin{equation}
    r(t,q)
    =
    \left(
    \{\tilde{\mathbf{O}}^{q}_{t-H},\ldots,\tilde{\mathbf{O}}^{q}_{t}\},
    \tilde{\mathbf{Y}}^{q}_{t},
    S,c,\ell,p
    \right),
    \qquad t\in\mathcal{A}.
\end{equation}
This construction keeps the anchor set fixed across clean and corrupted
settings while varying only the corruption source, family, severity, and
temporal placement.

\paragraph{Aggregation.}
For a corruption source \(S\in\{I,P,O\}\), \(CS^S\) denotes the clean score
under that source. \(RS^S_{\forall}\) averages all valid corruption families,
severities, and temporal protocols for source \(S\). \(RS^{\forall}_{\mathrm{Cur}}\),
\(RS^{\forall}_{\mathrm{Burst}}\), and \(RS^{\forall}_{\mathrm{Hist}}\) average over
sources for the corresponding temporal protocol. For the manual occupancy-state
source, \(RS^O_{\forall}\) is computed over the four non-traffic state
corruptions; Traffic is reported separately because it uses a fixed
mirrored-sequence protocol. The summary metrics used in the main paper are
\begin{equation}
    mRS=\frac{1}{3}\sum_{S\in\{I,P,O\}}RS^S_{\forall},\quad
    CRF_\beta=\frac{(1+\beta^2)CS^O\,mRS}{\beta^2CS^O+mRS},\quad
    UCR=\frac{CS^I+CS^P}{2CS^O}.
\end{equation}

\subsection{\texorpdfstring{\(CRF_{\beta}\)}{CRF-beta} Sensitivity}

\(CRF_{\beta}\) is used as a scale-preserving diagnostic for the clean-robust
trade-off, while \(mRS\) and the source-specific \(RS\) scores remain the
primary absolute robustness measures. The choice of \(\beta\) determines which
side of the trade-off receives more emphasis: \(CRF_2\), used in the main
tables, is robust-biased; \(CRF_1\) is the harmonic mean of clean and robust
performance; and \(CRF_{0.5}\) is clean-biased. Unlike retention ratios, all
three variants remain on the mIoU scale and penalize models that are strong in
only one of the two conditions.

\begin{table}[H]
    \centering
    \scriptsize
    \setlength{\tabcolsep}{4pt}
    \renewcommand{\arraystretch}{1.10}
    \caption{\(CRF_{\beta}\) sensitivity view of the 3D occupancy source table.
    The image and point-cloud variants use the source-local pairs
    \((CS,RS^{I}_{3D})\) and \((CS,RS^{P}_{3D})\), respectively. Scores are
    compared only within the same input-modality block.}
    \label{tab:app_crf_3d_sensitivity}
    \begin{adjustbox}{max width=\linewidth}
    \begin{tabular}{@{}lccccccc@{}}
        \toprule
        \multirow{2}{*}{\textbf{Method}} & \multirow{2}{*}{$\mathbf{CS}$}
        & \multicolumn{3}{c}{\textbf{Image Corruptions}}
        & \multicolumn{3}{c}{\textbf{Point-Cloud Corruptions}} \\
        \cmidrule(lr){3-5}\cmidrule(l){6-8}
        & & $\mathbf{RS}^{I}_{3D}$ & $\mathbf{CRF}^{I}_{1}$ &
        $\mathbf{CRF}^{I}_{0.5}$ & $\mathbf{RS}^{P}_{3D}$ &
        $\mathbf{CRF}^{P}_{1}$ & $\mathbf{CRF}^{P}_{0.5}$ \\
        \midrule
        \rowcolor{gray!10}\multicolumn{8}{l}{\textit{Camera-only methods}} \\
        SparseOcc & 21.7 & 9.7 & 13.41 & 17.40 & -- & -- & -- \\
        FB-OCC & 39.3 & 22.3 & 28.45 & 34.10 & -- & -- & -- \\
        ViewFormer & \underline{41.8} & 22.8 & 29.51 & 35.83 & -- & -- & -- \\
        COTR & 39.2 & 22.9 & 28.91 & 34.31 & -- & -- & -- \\
        PanoOcc & 36.6 & 25.0 & 29.71 & 33.49 & -- & -- & -- \\
        ALOcc & \textbf{44.0} & \underline{27.7} & \underline{34.00} & \textbf{39.37} & -- & -- & -- \\
        STCOcc & 41.7 & \textbf{30.2} & \textbf{35.03} & \underline{38.75} & -- & -- & -- \\
        \midrule
        \rowcolor{gray!10}\multicolumn{8}{l}{\textit{Camera--LiDAR methods}} \\
        OccFusion & 23.8 & 17.1 & 19.90 & 22.07 & 19.4 & 21.38 & 22.77 \\
        FusionOcc & \underline{45.4} & \underline{31.2} & \underline{36.98} &
        \underline{41.61} & \underline{42.2} & \underline{43.74} &
        \underline{44.72} \\
        SDGOcc & \textbf{51.6} & \textbf{35.4} & \textbf{41.99} &
        \textbf{47.27} & \textbf{42.6} & \textbf{46.67} & \textbf{49.51} \\
        \bottomrule
    \end{tabular}
    \end{adjustbox}
\end{table}

\begin{table}[H]
\centering\scriptsize
\setlength{\tabcolsep}{3pt}
\caption{$CRF_\beta$ sensitivity for the manual track on the same 4,519 anchors as the main tables. Traffic is excluded from $RS^O_{\forall}$.}
\label{tab:app_crf_manual_sensitivity}
\begin{adjustbox}{max width=\linewidth}
\begin{tabular}{@{}lccccc@{}}
\toprule
Method & $CS^O$ & $RS^O_{\forall}$ & $CRF_1$ & $CRF_{0.5}$ & Traffic \\
\midrule
OccWorld & 17.21 & 15.71 & 16.42 & 16.89 & 16.34 \\
DOME & 26.75 & 21.53 & 23.86 & 25.51 & 17.46 \\
COME & 34.52 & 31.87 & 33.14 & 33.95 & 19.12 \\
I$^2$-World & \underline{39.78} & \underline{36.24} & \underline{37.93} & \underline{39.02} & \textbf{38.07} \\
GenieDrive & \textbf{42.60} & \textbf{39.23} & \textbf{40.85} & \textbf{41.88} & \underline{22.75} \\
\bottomrule
\end{tabular}
\end{adjustbox}
\end{table}

The sensitivity tables illustrate the role of \(\beta\) without changing the
benchmark objective. Source selection in Table~\ref{tab:2_3d_occ_robustness}
is still governed by absolute corrupted occupancy quality, because the 4D
forecaster receives corrupted upstream states. The \(CRF_{\beta}\) variants
show how this choice would change under different clean-robust preferences.
For camera-only 3D occupancy, STCOcc has the strongest corrupted-state score
and \(CRF_1\), whereas ALOcc becomes highest under the clean-biased
\(CRF_{0.5}\) because of its stronger clean score. For camera--LiDAR 3D
occupancy, SDGOcc remains highest for both image and point-cloud corruptions
under all tested \(\beta\) values. In the five-model manual 4D track, GenieDrive
remains highest for \(CRF_1\) and \(CRF_{0.5}\), followed by I$^2$-World, COME,
DOME, and OccWorld.

\section{Detailed 3D Upstream Robustness}

\begin{table}[H]
    \centering
    \scriptsize
    \setlength{\tabcolsep}{8pt}
    \renewcommand{\arraystretch}{1.05}
    \caption{Full 3D image-corruption results by severity. Each row reports
    semantic mIoU under one severity level. Method blocks should be compared
    only within the same input-modality setting in the main paper.}
    \label{tab:app_3d_image_severity}
    
    \resizebox{\textwidth}{!}{%
    \begin{tabular}{@{}llcccccccc@{}}
        \toprule
        \textbf{Method} & \textbf{Severity} & Bright & Dark & Fog & Snow &
        Motion & Quant & Crash & Frame \\
        \midrule
        COTR & easy & 36.44 & 24.15 & 31.85 & 15.32 & \underline{36.77} & \underline{38.44} & 28.02 & 30.46 \\
             & mid  & 30.69 & 20.35 & 29.98 &  7.01 & 25.80 & 29.36 & 15.12 & 16.22 \\
             & hard & 27.26 & 14.48 & 27.61 &  5.74 & 21.18 & 16.53 & 13.26 &  9.41 \\
        \midrule
        ALOcc & easy & \textbf{40.36} & 29.47 & \textbf{36.86} & 16.71 & \textbf{39.58} & \textbf{41.15} & \textbf{36.09} & \textbf{38.75} \\
              & mid  & \textbf{34.39} & 25.10 & \textbf{35.24} &  6.62 & \underline{27.94} & \textbf{32.26} & \underline{26.58} & \underline{28.59} \\
              & hard & 30.36 & 18.06 & \textbf{33.34} &  5.68 & \underline{24.05} & 17.27 & \underline{20.61} & \underline{20.05} \\
        \midrule
        STCOcc & easy & \underline{36.92} & \textbf{33.32} & \underline{34.62} & \textbf{25.12} & 36.68 & 37.59 & \underline{34.18} & 34.62 \\
               & mid  & 33.70 & \textbf{31.34} & \underline{33.72} & \textbf{20.15} & \textbf{29.18} & \underline{31.92} & \textbf{31.30} & \textbf{29.63} \\
               & hard & \underline{31.60} & \textbf{28.11} & \underline{31.82} & \textbf{18.13} & \textbf{24.99} & \underline{22.64} & \textbf{28.28} & \textbf{26.76} \\
        \midrule
        FB-OCC & easy & 34.83 & 21.16 & 29.78 & 13.23 & 32.92 & 36.21 & 31.11 & 34.33 \\
               & mid  & 27.03 & 17.65 & 27.89 &  5.83 & 21.00 & 26.49 & 22.40 & 24.99 \\
               & hard & 22.13 & 12.08 & 25.21 &  4.93 & 17.45 & 13.09 & 17.36 & 17.08 \\
        \midrule
        ViewFormer & easy & 35.81 & 22.71 & 32.40 & 12.00 & 35.97 & 38.18 & 31.78 & \underline{36.24} \\
                   & mid  & 26.71 & 20.86 & 30.32 &  4.01 & 20.63 & 27.67 & 22.38 & 24.41 \\
                   & hard & 21.50 & 15.02 & 27.22 &  3.03 & 15.02 & 12.11 & 15.97 & 15.50 \\
        \midrule
        PanoOcc & easy & 35.96 & \underline{30.43} & 33.42 & \underline{19.55} & 33.00 & 35.82 & 28.16 & 29.22 \\
                & mid  & \underline{34.35} & \underline{26.16} & 32.31 & \underline{12.23} & 22.56 & 31.59 & 20.39 & 17.08 \\
                & hard & \textbf{32.53} & \underline{19.90} & 29.91 &  \underline{9.33} & 18.34 & \textbf{23.58} & 15.06 & 10.16 \\
        \midrule
        SparseOcc & easy & 16.18 & 11.92 & 13.66 &  4.53 & 13.99 & 15.60 & 12.21 & 12.91 \\
                  & mid  & 14.78 &  9.45 & 12.59 &  2.84 &  5.65 & 12.33 &  9.52 &  7.79 \\
                  & hard & 13.42 &  6.26 & 11.24 &  2.44 &  3.75 &  7.50 &  8.35 &  4.68 \\
        \midrule
        OccFusion & easy & 23.72 & 16.94 & 22.94 & 14.11 & 22.12 & 22.97 & 17.79 & 19.12 \\
                  & mid  & 22.99 & 13.64 & 22.52 & 10.93 & 15.99 & 19.91 & 14.64 & 13.28 \\
                  & hard & 22.40 & 10.21 & 21.84 & 10.79 & 13.65 & \underline{14.51} & 14.61 & 10.21 \\
        \midrule
        FusionOcc & easy & \underline{44.07} & \underline{29.93} & \underline{43.60} & \textbf{35.28} & \underline{41.39} & \underline{42.12} & \underline{32.35} & \textbf{37.43} \\
                  & mid  & \underline{40.29} & \underline{23.76} & \underline{42.37} & \textbf{23.79} & \underline{32.48} & \underline{30.74} & \textbf{23.26} & \textbf{24.71} \\
                  & hard & \underline{37.60} & \underline{16.87} & \underline{40.71} & \textbf{25.38} & \underline{28.40} & 14.37 & \textbf{21.48} & \textbf{16.48} \\
        \midrule
        SDGOcc & easy & \textbf{49.27} & \textbf{37.87} & \textbf{46.88} & \underline{32.84} & \textbf{49.24} & \textbf{50.53} & \textbf{34.25} & \underline{36.68} \\
               & mid  & \textbf{45.35} & \textbf{33.58} & \textbf{45.33} & \underline{21.23} & \textbf{40.89} & \textbf{44.82} & \underline{22.98} & \underline{21.64} \\
               & hard & \textbf{42.38} & \textbf{27.66} & \textbf{43.06} & \underline{20.31} & \textbf{36.51} & \textbf{32.39} & \underline{21.44} & \underline{14.66} \\
        \bottomrule
    \end{tabular}
    }
\end{table}

\begin{table}[H]
    \centering
    \scriptsize
    \setlength{\tabcolsep}{8pt}
    \renewcommand{\arraystretch}{1.05}
    \caption{Full 3D point-cloud-corruption results by severity for the
    camera--LiDAR baselines with completed point-cloud corruption runs.
    Cross denotes crosstalk and Sensor denotes cross-sensor corruption.}
\label{tab:app_3d_point_severity}
    
    \resizebox{\textwidth}{!}{%
    \begin{tabular}{@{}llcccccccc@{}}
        \toprule
        \textbf{Method} & \textbf{Severity} & Fog & Wet & Snow & Motion &
        Beam & Cross & Echo & Sensor \\
        \midrule
        OccFusion & light    & 21.90 & 23.36 & 21.99 & 22.62 & 21.44 &22.03 & 22.13 &17.31 \\
                  & moderate & 17.68 & 23.01 & 21.11 & 21.25 & 18.91 &19.86 & 21.23 &14.14 \\
                  & heavy    & 10.17 & 22.59 & 20.23 & 19.84 & 16.24 &17.84 & 19.17 &11.52 \\
        \midrule
        FusionOcc & light    & \underline{43.47} & \underline{45.68} & \underline{45.38} & \underline{40.73} & \underline{45.31} &\underline{43.37} & \underline{45.31} &\textbf{44.74} \\
                  & moderate & \textbf{39.09} & \underline{45.76} & \underline{44.45} & \underline{36.61} & \textbf{44.69} &\underline{41.42} & \underline{45.00} &\textbf{42.92} \\
                  & heavy    & \textbf{28.90} & \underline{45.76} & \underline{43.19} & \underline{33.13} & \textbf{43.36} &\underline{39.81} & \textbf{43.93} &\textbf{41.89} \\
        \midrule
        SDGOcc & light    & \textbf{47.30} & \textbf{50.87} & \textbf{49.45} & \textbf{43.99} & \textbf{48.21} &\textbf{49.53} & \textbf{49.08} &\underline{41.93} \\
               & moderate & \underline{36.48} & \textbf{50.34} & \textbf{47.90} & \textbf{39.07} & \underline{43.86} &\textbf{47.05} & \textbf{47.35} &\underline{34.68} \\
               & heavy    & \underline{12.07} & \textbf{49.48} & \textbf{45.98} & \textbf{34.83} & \underline{37.83} &\textbf{44.73} & \underline{42.29} &\underline{29.30} \\
        \bottomrule
    \end{tabular}
    }
\end{table}

\begin{table}[H]
    \centering
    \small
    \setlength{\tabcolsep}{6pt}
    \renewcommand{\arraystretch}{1.05}
    \caption{Additional 3D IoU-style metrics emitted by baseline-specific
    evaluators. These auxiliary metrics use the definitions of the original
    evaluators and are therefore reported for audit completeness rather than
    cross-method ranking. \(CS\) is the clean score and \(RS\) is averaged over
    the completed corruptions in the indicated source track.}
    \label{tab:app_3d_aux_iou}
    \begin{tabular}{@{}llccc@{}}
        \toprule
        \textbf{Method} & \textbf{Source} & \textbf{IoU metric} &
        $\mathbf{CS}$ & $\mathbf{RS}$ \\
        \midrule
        ALOcc & Image & Occupied IoU & 74.81 & 61.24 \\
        ViewFormer & Image & Geometric IoU & 71.29 & 54.56 \\
        SparseOcc & Image & RayIoU & 36.75 & 16.45 \\
        OccFusion & Image & Binary IoU & 36.44 & 34.04 \\
        OccFusion & Point cloud & Binary IoU & 36.44 & 30.79 \\
        \bottomrule
    \end{tabular}
\end{table}

\subsection{3D BEV Visualizations}

The following BEV visualizations show the 3D occupancy sources used by the
robustness pipeline before any 4D forecasting model is applied. All panels use
the same Occ3D BEV color map and fixed crop within each figure.

\begin{figure}[H]
    \centering
    \includegraphics[width=\textwidth]{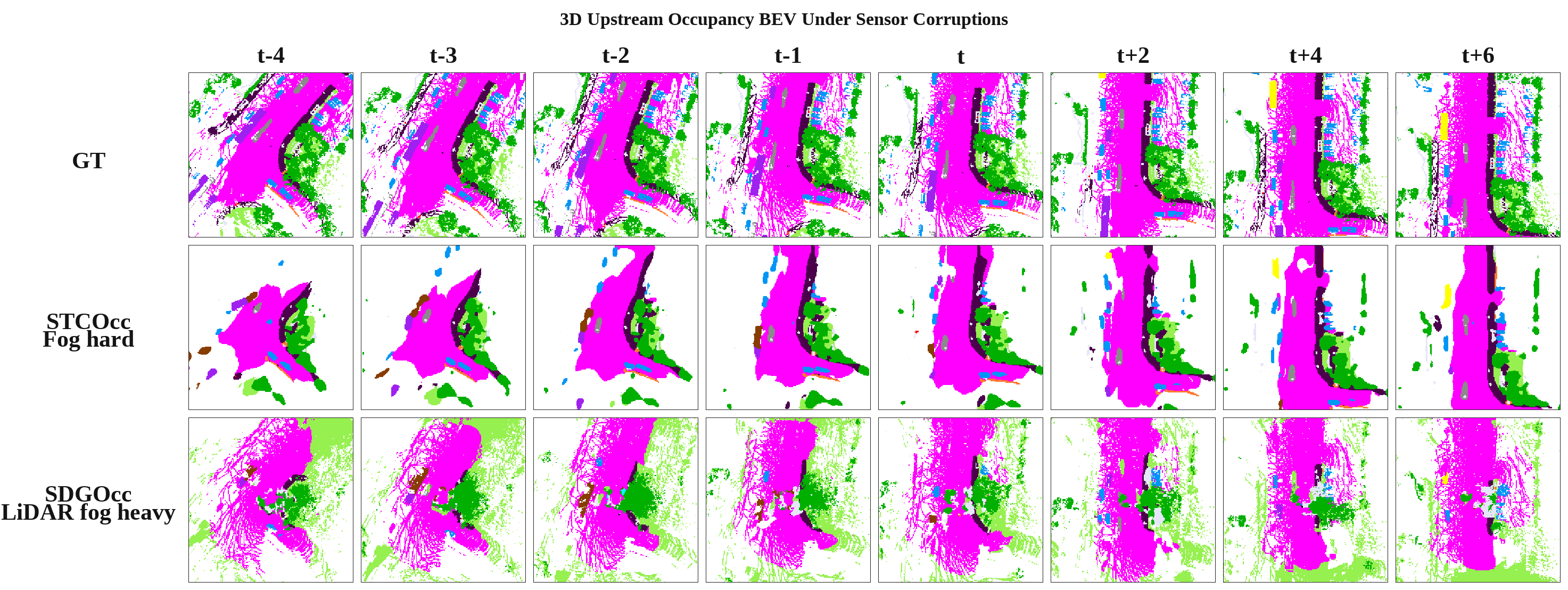}
    \caption{3D upstream BEV sequence for a shared scene. Rows show ground
    truth, STCOcc camera-only occupancy under hard fog, and SDGOcc
    point-cloud-fusion occupancy under heavy LiDAR fog. Columns cover the
    temporal window from \(t{-}4\) to \(t{+}6\). Only history and current states
    serve as forecasting inputs. Future columns are shown for visual reference.}
    \label{fig:app_3d_upstream_sequence}
\end{figure}

\begin{figure}[H]
    \centering
    \includegraphics[width=\textwidth]{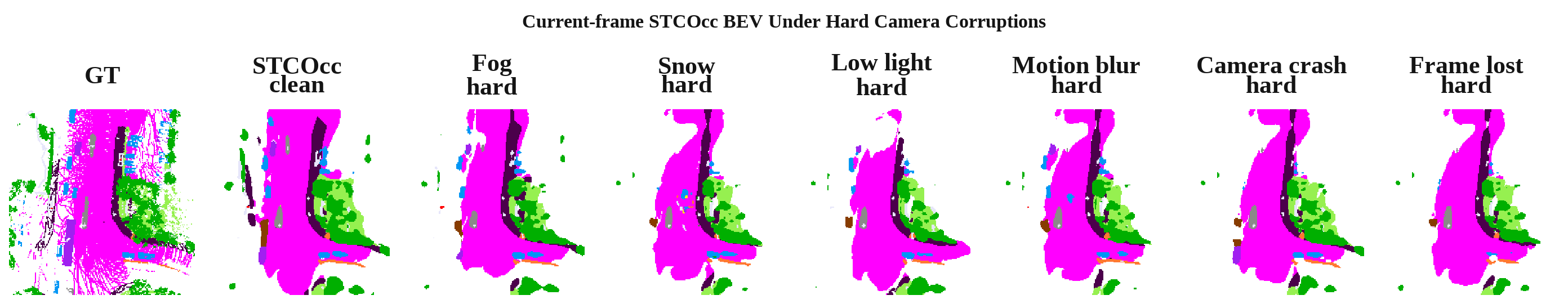}
    \caption{STCOcc camera-only BEV outputs for the same current frame under
    clean input and representative hard camera corruptions. The fixed crop
    makes source-side structural changes directly comparable across
    corruptions.}
    \label{fig:app_3d_camera_gallery}
\end{figure}

\begin{figure}[H]
    \centering
    \includegraphics[width=\textwidth]{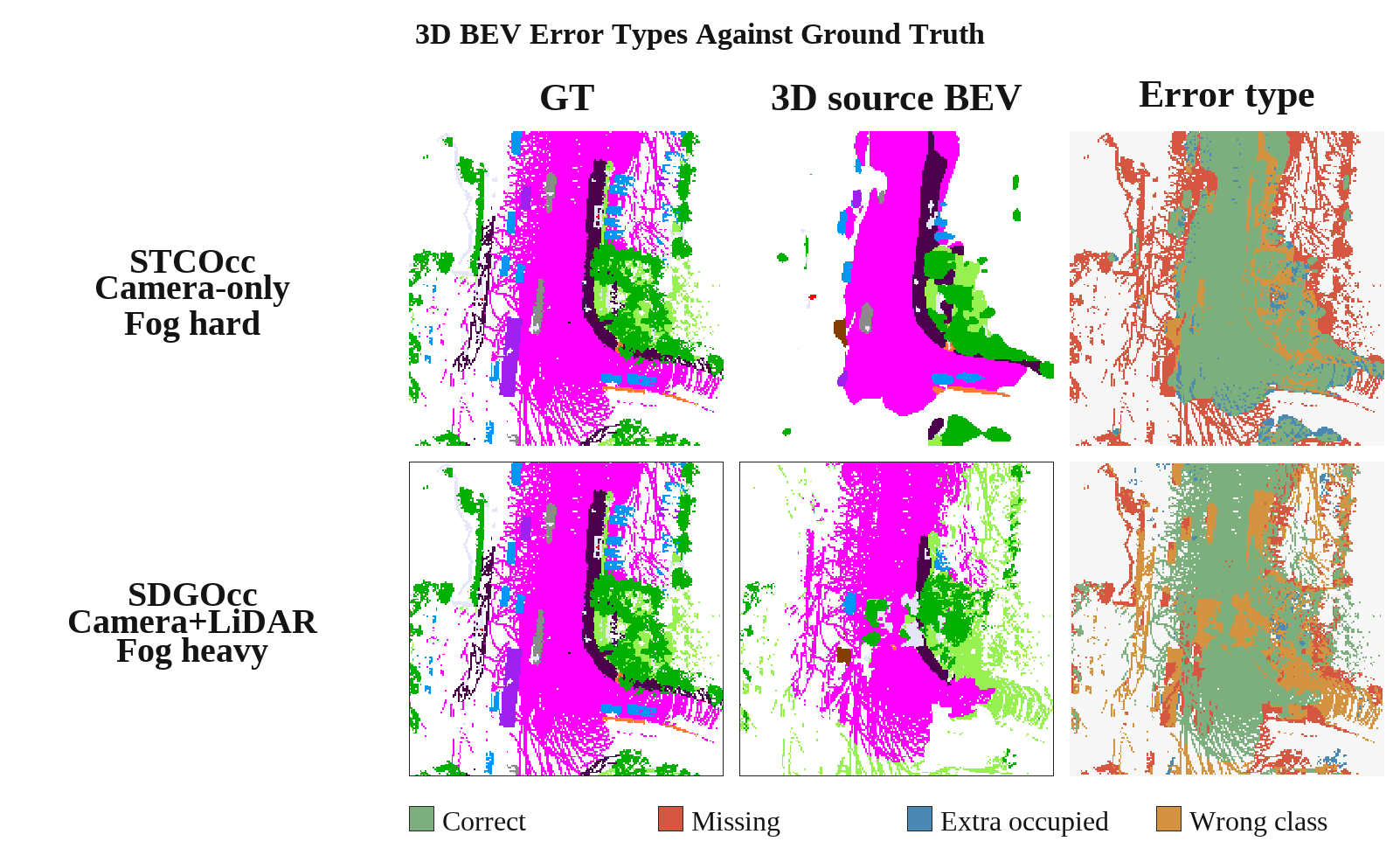}
    \caption{Current-frame BEV error maps against ground truth. Green denotes
    correctly predicted occupied cells, red missing occupied cells, blue
    hallucinated occupied cells, and orange semantic class swaps. The panels
    separate geometric failures from semantic confusion for STCOcc hard fog and
    SDGOcc heavy LiDAR fog.}
    \label{fig:app_3d_error_maps}
\end{figure}

\section{Detailed 4D Upstream-Corruption Results}

\begin{table}[H]
\centering\scriptsize
\setlength{\tabcolsep}{3pt}
\caption{4D semantic mIoU (\%) under STCOcc image corruption on the 4,519 OccStress-nuScenes anchors. Scores average 1s, 2s, and 3s. Family columns average three severities and protocols.}
\label{tab:app_4d_camera_upstream}
\begin{adjustbox}{max width=\linewidth}
\begin{tabular}{@{}lccccccccccccc@{}}
\toprule
Method & $CS^I$ & Bright & Dark & Fog & Snow & Motion & Quant & Crash & Frame & $RS^I_{\forall}$ & Cur & Burst & Hist \\
\midrule
OccWorld & 10.19 & 9.05 & 8.13 & 8.92 & 7.50 & 8.43 & 8.91 & 8.80 & 8.57 & 8.54 & 8.22 & 8.70 & 8.70 \\
DOME & 13.68 & 12.75 & 11.79 & 12.58 & 11.34 & 12.25 & 12.68 & 12.32 & 12.21 & 12.24 & 11.98 & 11.70 & 13.04 \\
COME & 15.58 & 14.41 & 13.31 & 14.24 & 12.64 & 13.69 & 14.29 & 13.80 & 13.61 & 13.75 & 13.36 & 13.22 & 14.67 \\
I$^2$-World & \underline{17.28} & \underline{16.03} & \underline{14.76} & \underline{15.80} & \underline{14.07} & \underline{15.27} & \underline{15.81} & \underline{15.45} & \underline{15.26} & \underline{15.31} & \underline{14.44} & \underline{14.50} & \underline{16.98} \\
GenieDrive & \textbf{17.79} & \textbf{16.52} & \textbf{15.26} & \textbf{16.27} & \textbf{14.62} & \textbf{15.70} & \textbf{16.26} & \textbf{16.02} & \textbf{15.80} & \textbf{15.81} & \textbf{14.97} & \textbf{14.97} & \textbf{17.49} \\
\bottomrule
\end{tabular}
\end{adjustbox}
\end{table}

\begin{table}[H]
\centering\scriptsize
\setlength{\tabcolsep}{3pt}
\caption{4D binary occupancy IoU (\%) under STCOcc image corruption on the 4,519 OccStress-nuScenes anchors. Scores average 1s, 2s, and 3s. Family columns average three severities and protocols.}
\label{tab:app_4d_camera_upstream_iou}
\begin{adjustbox}{max width=\linewidth}
\begin{tabular}{@{}lccccccccccccc@{}}
\toprule
Method & $CS^I$ & Bright & Dark & Fog & Snow & Motion & Quant & Crash & Frame & $RS^I_{\forall}$ & Cur & Burst & Hist \\
\midrule
OccWorld & 19.52 & 18.54 & 17.45 & 18.25 & 16.90 & 17.43 & 18.42 & 18.29 & 18.12 & 17.93 & 17.49 & 17.79 & 18.50 \\
DOME & 22.10 & 21.26 & 20.05 & 20.85 & 19.66 & 20.21 & 21.22 & 20.93 & 20.78 & 20.62 & 20.32 & 19.94 & 21.59 \\
COME & 24.30 & 23.34 & 21.99 & 22.85 & 21.46 & 22.02 & 23.20 & 22.84 & 22.63 & 22.54 & 22.24 & 21.84 & 23.54 \\
I$^2$-World & \underline{26.77} & \underline{25.61} & \underline{24.13} & \underline{25.16} & \underline{23.55} & \underline{24.14} & \underline{25.50} & \underline{25.19} & \underline{24.98} & \underline{24.78} & \underline{23.82} & \underline{23.83} & \underline{26.70} \\
GenieDrive & \textbf{27.37} & \textbf{26.22} & \textbf{24.65} & \textbf{25.70} & \textbf{24.15} & \textbf{24.73} & \textbf{26.10} & \textbf{25.79} & \textbf{25.55} & \textbf{25.36} & \textbf{24.46} & \textbf{24.36} & \textbf{27.26} \\
\bottomrule
\end{tabular}
\end{adjustbox}
\end{table}

\begin{table}[H]
\centering\scriptsize
\setlength{\tabcolsep}{3pt}
\caption{4D semantic mIoU (\%) under SDGOcc point-cloud corruption on the 4,519 OccStress-nuScenes anchors. Scores average 1s, 2s, and 3s. Family columns average three severities and protocols. Cross denotes crosstalk and Sensor denotes cross-sensor corruption.}
\label{tab:app_4d_point_upstream}
\begin{adjustbox}{max width=\linewidth}
\begin{tabular}{@{}lccccccccccccc@{}}
\toprule
Method & $CS^P$ & Fog & Wet & Snow & Motion & Beam & Cross & Echo & Sensor & $RS^P_{\forall}$ & Cur & Burst & Hist \\
\midrule
OccWorld & 11.03 & 7.57 & 10.83 & 10.65 & 9.41 & 9.95 & 10.55 & 10.42 & 9.07 & 9.81 & 9.60 & 9.58 & 10.24 \\
DOME & 15.73 & 12.12 & 15.54 & 15.27 & 14.30 & 14.45 & 15.22 & 14.80 & 13.21 & 14.37 & 14.16 & 13.86 & 15.08 \\
COME & 19.56 & 15.55 & 19.34 & 19.13 & 17.66 & 17.91 & 19.07 & 18.50 & 16.25 & 17.93 & 17.58 & 17.20 & 19.00 \\
I$^2$-World & \underline{21.44} & \underline{16.59} & \underline{21.05} & \underline{20.72} & \underline{18.51} & \underline{19.40} & \underline{20.57} & \underline{20.20} & \underline{17.58} & \underline{19.33} & \underline{18.36} & \underline{18.34} & \underline{21.29} \\
GenieDrive & \textbf{23.37} & \textbf{18.27} & \textbf{22.96} & \textbf{22.65} & \textbf{20.41} & \textbf{21.23} & \textbf{22.52} & \textbf{22.15} & \textbf{19.23} & \textbf{21.18} & \textbf{20.23} & \textbf{20.13} & \textbf{23.17} \\
\bottomrule
\end{tabular}
\end{adjustbox}
\end{table}

\begin{table}[H]
\centering\scriptsize
\setlength{\tabcolsep}{3pt}
\caption{4D binary occupancy IoU (\%) under SDGOcc point-cloud corruption on the 4,519 OccStress-nuScenes anchors. Scores average 1s, 2s, and 3s. Family columns average three severities and protocols. Cross denotes crosstalk and Sensor denotes cross-sensor corruption.}
\label{tab:app_4d_point_upstream_iou}
\begin{adjustbox}{max width=\linewidth}
\begin{tabular}{@{}lccccccccccccc@{}}
\toprule
Method & $CS^P$ & Fog & Wet & Snow & Motion & Beam & Cross & Echo & Sensor & $RS^P_{\forall}$ & Cur & Burst & Hist \\
\midrule
OccWorld & 21.73 & 18.28 & 21.59 & 21.40 & 20.61 & 21.01 & 21.27 & 21.63 & 20.30 & 20.76 & 20.62 & 20.64 & 21.02 \\
DOME & 27.30 & 23.61 & 27.14 & 26.84 & 26.43 & 26.44 & 26.71 & 27.20 & 25.42 & 26.22 & 26.07 & 25.84 & 26.76 \\
COME & 32.78 & 28.61 & 32.57 & 32.36 & 31.60 & 31.64 & 32.24 & 32.63 & 30.31 & 31.49 & 31.30 & 31.01 & 32.18 \\
I$^2$-World & \underline{36.21} & \underline{31.43} & \underline{35.98} & \underline{35.53} & \underline{34.43} & \underline{35.05} & \underline{35.30} & \underline{36.13} & \underline{33.71} & \underline{34.69} & \underline{33.96} & \underline{33.96} & \underline{36.16} \\
GenieDrive & \textbf{38.50} & \textbf{33.54} & \textbf{38.27} & \textbf{37.80} & \textbf{36.80} & \textbf{37.27} & \textbf{37.54} & \textbf{38.40} & \textbf{35.88} & \textbf{36.94} & \textbf{36.24} & \textbf{36.17} & \textbf{38.40} \\
\bottomrule
\end{tabular}
\end{adjustbox}
\end{table}

\subsection{Additional Upstream Sources}
\label{sec:app_additional_upstream}

We also evaluate ALOcc and FusionOcc-BaseWoMask as upstream sources for
I$^2$-World and GenieDrive on the same 4,519 OccStress-nuScenes anchors.
Each ALOcc chain covers 73 image-source configurations. Each FusionOcc chain
covers 145 configurations, comprising one shared clean condition and 72
corrupted conditions per source. Unmodified input positions use the same
estimator's clean predictions. Table~\ref{tab:app_additional_upstream} reports
each source separately, without redefining the main benchmark's source set.

\begin{table}[H]
\centering\small
\setlength{\tabcolsep}{4pt}
\caption{Additional 4D upstream-chain results on OccStress-nuScenes. Semantic
mIoU (\%) averages the physical 1s, 2s, and 3s horizons. $S$ identifies the
corrupted sensor source, not the estimator's full input modality. $RS^S_{\forall}$
averages all 72 corrupted conditions, and each protocol column averages
eight families and three severities.}
\label{tab:app_additional_upstream}
\begin{tabular}{@{}llcrrrrr@{}}
\toprule
Forecaster & Upstream & $S$ & $CS^S$ & $RS^S_{\forall}$ & $RS^S_{\rm Cur}$ & $RS^S_{\rm Burst}$ & $RS^S_{\rm Hist}$ \\
\midrule
I$^2$-World & ALOcc & I & 9.29 & 7.21 & 6.27 & 6.28 & 9.07 \\
GenieDrive & ALOcc & I & 8.86 & 7.14 & 6.34 & 6.26 & 8.81 \\
I$^2$-World & FusionOcc & I & 10.42 & 8.91 & 8.18 & 8.17 & 10.37 \\
I$^2$-World & FusionOcc & P & 10.42 & 9.93 & 9.70 & 9.70 & 10.38 \\
GenieDrive & FusionOcc & I & 9.52 & 8.16 & 7.54 & 7.47 & 9.46 \\
GenieDrive & FusionOcc & P & 9.52 & 9.14 & 8.97 & 8.94 & 9.49 \\
\bottomrule
\end{tabular}
\end{table}

\section{One-Stage Camera-Based Forecasting}
\label{sec:app_sparseworld}

SparseWorld-TC~\cite{dang2025sparseworld} directly forecasts future occupancy
from camera sequences and completes all 73 applicable camera-source
configurations per dataset, using the nuScenes checkpoint zero-shot on OccStress-Waymo
and OccStress-CARLA (Table~\ref{tab:app_sparseworld}). The occupancy-state and point-source
tracks are inapplicable, so we report only source-specific scores and omit
$UCR$, $mRS$, and $CRF_2$.

\begin{table}[H]
\centering\small
\setlength{\tabcolsep}{6pt}
\caption{SparseWorld-TC camera-source robustness on the common H4/F6
inventories: 4,519 OccStress-nuScenes, 5,978 OccStress-Waymo, and 330 OccStress-CARLA anchors. Semantic
mIoU (\%) averages 1s, 2s, and 3s. $RS^I_{\forall}$ averages 72 corrupted
configurations, and each protocol column averages eight families and three
severities.}
\label{tab:app_sparseworld}
\begin{tabular}{@{}lrrrrr@{}}
\toprule
Dataset & $CS^I$ & $RS^I_{\forall}$ & $RS^I_{\rm Cur}$ & $RS^I_{\rm Burst}$ & $RS^I_{\rm Hist}$ \\
\midrule
OccStress-nuScenes & 13.18 & 11.74 & 12.44 & 11.79 & 10.99 \\
OccStress-Waymo & 3.69 & 3.00 & 2.70 & 2.66 & 3.64 \\
OccStress-CARLA & 4.59 & 3.57 & 3.06 & 3.05 & 4.59 \\
\bottomrule
\end{tabular}
\end{table}

\section{Detailed Manual Occupancy-State Corruptions}

\begin{table}[H]
\centering\scriptsize
\setlength{\tabcolsep}{3pt}
\caption{Manual corruption semantic mIoU (\%) on the 4,519 OccStress-nuScenes anchors. Each entry averages three severities and the 1s, 2s, and 3s horizons. Rankings are within each protocol block.}
\label{tab:app_manual_protocol}
\begin{adjustbox}{max width=\linewidth}
\begin{tabular}{@{}lccccc@{}}
\toprule
Method & Semantic & Hole & Dropout & Misaligned & $RS^O_p$ \\
\midrule
\multicolumn{6}{l}{\textit{Cur}} \\
OccWorld & 13.73 & 16.93 & 13.76 & 17.21 & 15.41 \\
DOME & 21.53 & 26.61 & 22.41 & 11.49 & 20.51 \\
COME & 26.99 & 34.24 & 28.86 & 34.52 & 31.15 \\
I$^2$-World & \underline{29.51} & \underline{39.14} & \underline{31.88} & \underline{37.76} & \underline{34.57} \\
GenieDrive & \textbf{32.04} & \textbf{41.70} & \textbf{34.85} & \textbf{42.42} & \textbf{37.75} \\

\multicolumn{6}{l}{\textit{Burst}} \\
OccWorld & 13.29 & 16.83 & 13.10 & 17.21 & 15.11 \\
DOME & 20.57 & 26.55 & 21.28 & 10.74 & 19.78 \\
COME & 26.52 & 34.11 & 28.03 & 34.52 & 30.80 \\
I$^2$-World & \underline{29.50} & \underline{39.11} & \underline{31.76} & \underline{37.76} & \underline{34.53} \\
GenieDrive & \textbf{31.93} & \textbf{41.65} & \textbf{34.57} & \textbf{42.29} & \textbf{37.61} \\

\multicolumn{6}{l}{\textit{Hist}} \\
OccWorld & 16.70 & 17.10 & 15.40 & 17.21 & 16.60 \\
DOME & 25.46 & 26.70 & 24.83 & 20.19 & 24.30 \\
COME & 33.50 & 34.46 & 32.21 & 34.52 & 33.67 \\
I$^2$-World & \underline{39.67} & \underline{39.77} & \underline{39.25} & \underline{39.78} & \underline{39.62} \\
GenieDrive & \textbf{42.36} & \textbf{42.55} & \textbf{41.99} & \textbf{42.40} & \textbf{42.33} \\
\bottomrule
\end{tabular}
\end{adjustbox}
\end{table}

\begin{table}[H]
\centering\scriptsize
\setlength{\tabcolsep}{3pt}
\caption{Manual corruption binary occupancy IoU (\%) on the 4,519 OccStress-nuScenes anchors. Each entry averages three severities and the 1s, 2s, and 3s horizons. Rankings are within each protocol block.}
\label{tab:app_manual_protocol_iou}
\begin{adjustbox}{max width=\linewidth}
\begin{tabular}{@{}lccccc@{}}
\toprule
Method & Semantic & Hole & Dropout & Misaligned & $RS^O_p$ \\
\midrule
\multicolumn{6}{l}{\textit{Cur}} \\
OccWorld & 26.00 & 26.58 & 22.21 & 26.64 & 25.36 \\
DOME & 35.96 & 36.39 & 31.56 & 21.06 & 31.24 \\
COME & 44.44 & 45.04 & 38.58 & 45.11 & 43.29 \\
I$^2$-World & \underline{48.80} & \underline{49.97} & \underline{41.50} & \underline{49.64} & \underline{47.48} \\
GenieDrive & \textbf{50.94} & \textbf{51.59} & \textbf{43.43} & \textbf{51.79} & \textbf{49.44} \\

\multicolumn{6}{l}{\textit{Burst}} \\
OccWorld & 26.01 & 26.55 & 21.75 & 26.64 & 25.24 \\
DOME & 35.89 & 36.36 & 29.98 & 20.41 & 30.66 \\
COME & 44.39 & 45.00 & 37.29 & 45.11 & 42.95 \\
I$^2$-World & \underline{48.79} & \underline{49.96} & \underline{41.47} & \underline{49.64} & \underline{47.46} \\
GenieDrive & \textbf{50.94} & \textbf{51.57} & \textbf{43.22} & \textbf{51.75} & \textbf{49.37} \\

\multicolumn{6}{l}{\textit{Hist}} \\
OccWorld & 26.55 & 26.63 & 25.53 & 26.64 & 26.33 \\
DOME & 36.23 & 36.42 & 34.73 & 30.10 & 34.37 \\
COME & 44.90 & 45.10 & 41.99 & 45.11 & 44.28 \\
I$^2$-World & \underline{50.03} & \underline{50.13} & \underline{49.93} & \underline{50.14} & \underline{50.06} \\
GenieDrive & \textbf{51.81} & \textbf{51.81} & \textbf{51.54} & \textbf{51.79} & \textbf{51.74} \\
\bottomrule
\end{tabular}
\end{adjustbox}
\end{table}

\begin{table}[H]
\centering\scriptsize
\setlength{\tabcolsep}{3pt}
\caption{Traffic mirroring on the 4,519 OccStress-nuScenes anchors. Inputs, targets, and motion metadata use the same mirrored coordinates. Scores average 1s, 2s, and 3s.}
\label{tab:app_manual_traffic}
\begin{adjustbox}{max width=\linewidth}
\begin{tabular}{@{}lcccccc@{}}
\toprule
Method & $CS^O$ & Traffic & $\Delta$ & Clean IoU & Traffic IoU & $\Delta$ \\
\midrule
OccWorld & 17.21 & 16.34 & \textbf{-0.87} & 26.64 & 25.92 & \textbf{-0.72} \\
DOME & 26.75 & 17.46 & -9.28 & 36.44 & 26.02 & -10.42 \\
COME & 34.52 & 19.12 & -15.40 & 45.11 & 27.66 & -17.45 \\
I$^2$-World & \underline{39.78} & \textbf{38.07} & \underline{-1.71} & \underline{50.14} & \textbf{49.26} & \underline{-0.88} \\
GenieDrive & \textbf{42.60} & \underline{22.75} & -19.85 & \textbf{51.82} & \underline{32.13} & -19.69 \\
\bottomrule
\end{tabular}
\end{adjustbox}
\end{table}

\subsection{Corruption-Aware Fine-Tuning}
\label{sec:app_mitigation}

We fine-tune I$^2$-World on a 60/40 mixture of clean and corrupted nuScenes
training clips, balanced across semantic, dropout, hole, and misalignment
settings. Future targets remain clean, and traffic mirroring is held out.
Starting from the released checkpoint, we use AdamW with learning rate
\(5\times10^{-5}\), weight decay 0.01, and 1,200 iterations, with a
clean-sample distillation loss of weight 0.5. Table~\ref{tab:app_mitigation}
reports the final EMA checkpoint on the three content-corruption families
and held-out traffic. This run does not improve their mean scores, so it
provides a negative baseline rather than evidence of a general mitigation.

\begin{table}[H]
\centering\small
\setlength{\tabcolsep}{8pt}
\caption{Corruption-aware fine-tuning of I$^2$-World on 4,519 OccStress-nuScenes
anchors. Semantic mIoU (\%) averages 1s, 2s, and 3s. Content-corruption rows
also average three severities and three protocols. Traffic uses its single
mirrored-sequence protocol. $\Delta$ is fine-tuned minus original, computed
before rounding.}
\label{tab:app_mitigation}
\begin{tabular}{@{}lrrr@{}}
\toprule
Setting & Original & Fine-tuned & $\Delta$ \\
\midrule
Semantic & 32.90 & 32.84 & -0.05 \\
Dropout & 34.29 & 33.25 & -1.05 \\
Hole & 39.34 & 38.08 & -1.26 \\
Traffic (held out) & 38.07 & 37.77 & -0.30 \\
\bottomrule
\end{tabular}
\end{table}

\section{Class-Wise Clean Forecasting Diagnostics}

The following tables expand the clean occupancy-state track at the semantic
class level. Per-class entries are semantic IoU, while the companion summary
reports semantic mIoU and binary occupancy IoU at the same 1s, 2s, and 3s
forecasting horizons used by the main tables.

\begin{table}[H]
    \centering
    \scriptsize
    \setlength{\tabcolsep}{1.8pt}
    \renewcommand{\arraystretch}{1.12}
    \caption{Clean 4D forecasting by semantic class. Class entries are
    per-class semantic IoU (\%) averaged over the 1s, 2s, and 3s forecasting
    horizons on the 4,519-anchor population. mIoU averages the 17 occupied
    classes, excluding free space. The colored squares follow the
    OccWorld Occ3D BEV palette.}
    \label{tab:app_clean_class_iou}
    \resizebox{\textwidth}{!}{%
    \begin{tabular}{@{}l*{18}{c}@{}}
        \toprule
        \textbf{Method}
        & \occhead{occOthers}{others}
        & \occhead{occBarrier}{barrier}
        & \occhead{occBicycle}{bicycle}
        & \occhead{occBus}{bus}
        & \occhead{occCar}{car}
        & \occhead{occConstruction}{constr. veh.}
        & \occhead{occMotorcycle}{motor.}
        & \occhead{occPedestrian}{ped.}
        & \occhead{occTrafficCone}{traffic cone}
        & \occhead{occTrailer}{trailer}
        & \occhead{occTruck}{truck}
        & \occhead{occDriveable}{drive. surf.}
        & \occhead{occOtherFlat}{other flat}
        & \occhead{occSidewalk}{sidewalk}
        & \occhead{occTerrain}{terrain}
        & \occhead{occManmade}{manmade}
        & \occhead{occVegetation}{veget.}
        & \textbf{mIoU}
        \\
        \midrule
OccWorld & 12.3 & 21.1 & 8.4 & 20.8 & 19.9 & 12.9 & 7.8 & 8.9 & 8.5 & 13.2 & 17.7 & 35.1 & 24.0 & 24.1 & 20.7 & 17.2 & 20.1 & 17.21 \\
DOME & 27.0 & 32.9 & 17.6 & 25.0 & 25.1 & 32.0 & 16.1 & 9.4 & 17.3 & 23.7 & 27.1 & 42.7 & 33.7 & 32.5 & 29.3 & 29.1 & 34.1 & 26.75 \\
COME & 35.3 & 42.6 & 24.3 & 30.5 & 30.8 & 35.1 & 27.0 & 20.2 & 26.1 & 29.6 & 28.0 & 53.3 & 43.8 & 41.5 & 40.3 & 38.0 & 40.3 & 34.52 \\
I$^2$-World & 40.9 & 50.2 & 26.1 & 35.0 & 36.0 & 39.1 & 26.8 & 18.6 & 31.2 & 36.4 & 40.7 & 62.3 & 50.7 & 49.7 & 50.3 & 42.4 & 39.9 & 39.78 \\
GenieDrive & 46.1 & 51.7 & 29.9 & 36.5 & 38.5 & 43.9 & 24.2 & 21.3 & 36.8 & 39.5 & 41.3 & 66.1 & 55.0 & 55.3 & 53.1 & 43.4 & 41.5 & 42.60 \\
        \bottomrule
    \end{tabular}
    }
\end{table}

\begin{table}[H]
\centering\scriptsize
\setlength{\tabcolsep}{3pt}
\caption{Clean forecasting on the 4,519 OccStress-nuScenes anchors. Avg averages physical 1s, 2s, and 3s horizons.}
\label{tab:app_clean_horizon_miou_iou}
\begin{adjustbox}{max width=\linewidth}
\begin{tabular}{@{}lcccccccc@{}}
\toprule
Method & mIoU 1s & 2s & 3s & Avg & IoU 1s & 2s & 3s & Avg \\
\midrule
OccWorld & 25.82 & 15.22 & 10.61 & 17.21 & 34.64 & 25.08 & 20.20 & 26.64 \\
DOME & 34.37 & 25.71 & 20.15 & 26.75 & 43.12 & 35.77 & 30.43 & 36.44 \\
COME & 42.58 & 33.47 & 27.51 & 34.52 & 51.04 & 44.64 & 39.67 & 45.11 \\
I$^2$-World & \underline{47.95} & \underline{38.59} & \underline{32.79} & \underline{39.78} & \underline{54.76} & \underline{49.75} & \underline{45.90} & \underline{50.14} \\
GenieDrive & \textbf{50.47} & \textbf{41.49} & \textbf{35.84} & \textbf{42.60} & \textbf{56.88} & \textbf{51.48} & \textbf{47.10} & \textbf{51.82} \\
\bottomrule
\end{tabular}
\end{adjustbox}
\end{table}

\section{Horizon-Wise Diagnostics}

\begin{figure}[H]
\centering
\includegraphics[width=\linewidth]{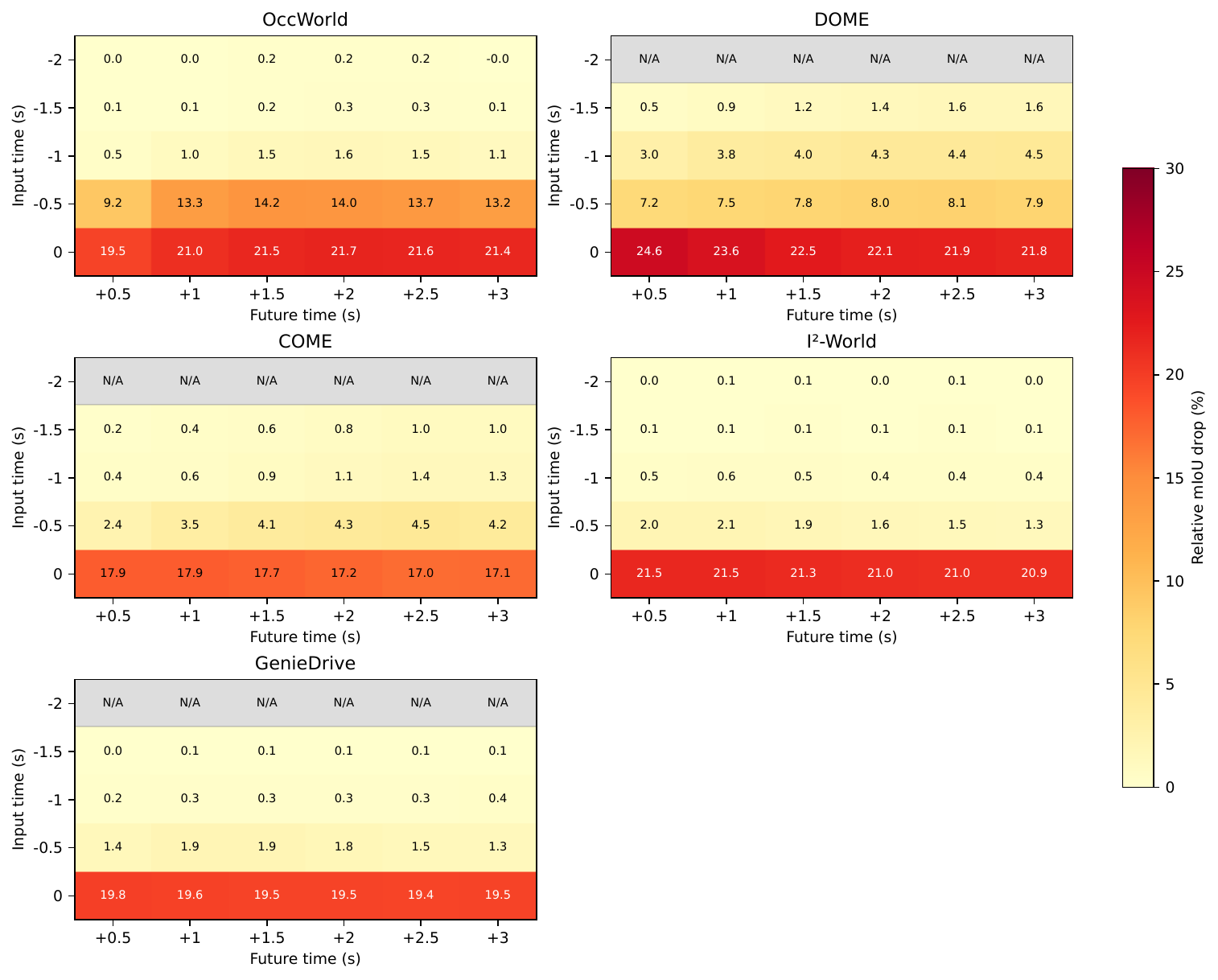}
\caption{Position-wise relative mIoU drop (\%) averaged over ten hard settings.
Rows within each panel are input times and columns are future times, in seconds
relative to the anchor. N/A means that the model does not observe that state.}
\label{fig:app_position_sweep_heatmaps}
\end{figure}

Figure~\ref{fig:app_position_sweep_heatmaps} expands the five-model position
analysis over the same ten hard settings and 4,519 OccStress-nuScenes anchors.
The settings are manual dropout, hole, and misalignment, and STCOcc brightness,
camera crash, color quantization, fog, frame loss, low light, and motion blur.
Each cell averages relative mIoU drops over these settings at one physical
forecasting horizon. Five input positions are shown for OccWorld and I$^2$-World.
The earliest position is marked N/A for COME, DOME, and GenieDrive because it
is outside their input window. Cross-model averages use only the four shared
positions. The separate three-setting I$^2$-World bootstrap contrast below
retains its original full five-position definition.

\section{Statistical Uncertainty and Validity Audits}
\label{sec:app_uncertainty}

\subsection{Scene-Clustered Paired Bootstrap}

Adjacent temporal anchors overlap and are therefore correlated. We use the
nuScenes validation scene as the resampling unit rather than treating anchors
as independent. For I$^2$-World, the analysis covers all 184 main protocols on
the common 4,519-anchor population within 150 scenes, together with 18 fixed
position-sweep protocols. We perform 10,000 paired scene-clustered bootstrap
resamples with seed 20260724. Every replicate uses the same sampled-scene
multiplicities for clean and corrupted conditions. Per-scene confusion counts
are pooled before computing semantic mIoU at the physical 1s, 2s, and 3s
forecasting horizons. We report percentile 95\% confidence intervals.

For the position check, let \(D_j\) denote the mIoU degradation when only input
position \(j\) is corrupted. The contrast
\(D_t-\overline{D}_{t-4:t-1}\) averages three fixed settings: manual
semantic-hard, STCOcc snow-hard, and SDGOcc snow-heavy. Table~\ref{tab:app_uncertainty}
reports the aggregate intervals and the paired claim checks used in the main
analysis.

\begin{table}[H]
    \centering
    \small
    \setlength{\tabcolsep}{7pt}
    \renewcommand{\arraystretch}{1.10}
    \caption{Scene-clustered uncertainty estimates for I$^2$-World on
    OccStress-nuScenes. Differences are computed within each paired bootstrap
    replicate.}
    \label{tab:app_uncertainty}
    \begin{tabular}{@{}lcc@{}}
        \toprule
        \textbf{Quantity} & \textbf{Estimate} & \textbf{95\% CI} \\
        \midrule
        \(mRS\) & 23.63 & [22.92, 24.30] \\
        \(CRF_2\) & 25.71 & [24.97, 26.43] \\
        \(CS^O-mRS\) & 16.15 & [15.50, 16.81] \\
        \(CS^I-RS^I_{\forall}\) & 1.98 & [1.85, 2.11] \\
        \(CS^P-RS^P_{\forall}\) & 2.11 & [1.99, 2.23] \\
        \(CS^O-RS^O_{\forall}\) & 3.54 & [3.34, 3.74] \\
        \(RS^P_{\forall}-RS^I_{\forall}\) & 4.02 & [3.12, 4.97] \\
        \(D_t-\overline{D}_{t-4:t-1}\) & 7.08 & [6.74, 7.42] \\
        \bottomrule
    \end{tabular}
\end{table}

\paragraph{Source-, family-, and configuration-level intervals.}
Tables~\ref{tab:app_uncertainty_sources}--\ref{tab:app_uncertainty_protocols}
give the full I$^2$-World semantic mIoU breakdown at 1s, 2s, and 3s.
Brackets denote marginal 95\% intervals. Aggregates are recomputed within
each paired replicate. Protocol labels follow the actual corrupted input
positions, and robust means exclude traffic.

\begin{table}[H]
\centering\scriptsize
\setlength{\tabcolsep}{3pt}
\caption{Source-wise and source-conditioned protocol scores with 95\% CIs.
Sources $I/P/O$ use STCOcc, SDGOcc, and ground-truth occupancy, respectively.}
\label{tab:app_uncertainty_sources}
\begin{tabular}{@{}lccccc@{}}
\toprule
$S$ & $CS^S$ & $RS^S_{\forall}$ & $RS^S_{\rm Cur}$ & $RS^S_{\rm Burst}$ & $RS^S_{\rm Hist}$ \\
\midrule
I & 17.28 [16.54, 18.04] & 15.31 [14.63, 15.99] & 14.44 [13.80, 15.10] & 14.50 [13.85, 15.16] & 16.98 [16.25, 17.73] \\
P & 21.44 [20.43, 22.47] & 19.33 [18.43, 20.26] & 18.36 [17.50, 19.25] & 18.34 [17.48, 19.23] & 21.29 [20.29, 22.31] \\
O & 39.78 [38.66, 40.87] & 36.24 [35.23, 37.22] & 34.57 [33.61, 35.51] & 34.53 [33.57, 35.47] & 39.62 [38.49, 40.71] \\
\bottomrule
\end{tabular}
\end{table}

\begin{table}[H]
\centering\footnotesize
\setlength{\tabcolsep}{10pt}
\caption{Corruption-family scores with 95\% CIs. Each ordinary family
averages nine severity--protocol configurations. Traffic has one configuration.}
\label{tab:app_uncertainty_families}
\begin{tabular}{@{}llc@{}}
\toprule
$S$ & Family & mIoU [95\% CI] \\
\midrule
I & Brightness & 16.03 [15.32, 16.74] \\
I & Crash & 15.45 [14.76, 16.17] \\
I & Quant & 15.81 [15.11, 16.53] \\
I & Fog & 15.80 [15.13, 16.49] \\
I & Frame & 15.26 [14.59, 15.95] \\
I & Dark & 14.76 [14.10, 15.43] \\
I & Motion & 15.27 [14.60, 15.96] \\
I & Snow & 14.07 [13.41, 14.74] \\
P & Beam & 19.40 [18.49, 20.35] \\
P & Sensor & 17.58 [16.71, 18.48] \\
P & Cross & 20.57 [19.61, 21.56] \\
P & Fog & 16.59 [15.78, 17.41] \\
P & Echo & 20.20 [19.25, 21.19] \\
P & Motion & 18.51 [17.69, 19.35] \\
P & Snow & 20.72 [19.74, 21.72] \\
P & Wet & 21.05 [20.07, 22.06] \\
O & Dropout & 34.29 [33.29, 35.31] \\
O & Hole & 39.34 [38.21, 40.45] \\
O & Misalignment & 38.43 [37.32, 39.49] \\
O & Semantic & 32.90 [31.99, 33.74] \\
O & Traffic & 38.07 [37.01, 39.08] \\
\bottomrule
\end{tabular}
\end{table}

\begingroup
\scriptsize
\setlength{\tabcolsep}{4pt}
\begin{longtable}{@{}lllccc@{}}
\caption{All corrupted-configuration scores with 95\% CIs. Together with
the three clean scores in Table~\ref{tab:app_uncertainty_sources}, these cover
all 184 configurations. E/M/H denotes increasing severity. Traffic's single
fixed-protocol result is shown across the three protocol columns.}
\label{tab:app_uncertainty_protocols}\\
\toprule
$S$ & Family & Severity & Cur & Burst & Hist \\
\midrule
\endfirsthead
\multicolumn{6}{l}{\tablename~\thetable{} (continued)}\\
\toprule
$S$ & Family & Severity & Cur & Burst & Hist \\
\midrule
\endhead
\midrule
\multicolumn{6}{r}{Continued on next page}\\
\endfoot
\bottomrule
\endlastfoot
I & Brightness & E & 16.65 [15.91, 17.39] & 16.66 [15.92, 17.40] & 17.23 [16.49, 18.00] \\
I & Brightness & M & 15.31 [14.59, 16.02] & 15.39 [14.66, 16.10] & 17.14 [16.39, 17.89] \\
I & Brightness & H & 14.36 [13.65, 15.04] & 14.43 [13.72, 15.11] & 17.09 [16.35, 17.85] \\
I & Crash & E & 16.07 [15.32, 16.81] & 16.09 [15.35, 16.84] & 17.08 [16.33, 17.84] \\
I & Crash & M & 15.11 [14.37, 15.85] & 15.16 [14.41, 15.91] & 16.96 [16.22, 17.72] \\
I & Crash & H & 12.98 [12.32, 13.66] & 13.01 [12.35, 13.69] & 16.63 [15.90, 17.38] \\
I & Quant & E & 17.01 [16.27, 17.75] & 17.01 [16.27, 17.75] & 17.23 [16.48, 17.99] \\
I & Quant & M & 15.52 [14.81, 16.23] & 15.56 [14.85, 16.29] & 17.16 [16.43, 17.91] \\
I & Quant & H & 12.93 [12.28, 13.60] & 13.10 [12.43, 13.77] & 16.82 [16.09, 17.57] \\
I & Fog & E & 15.49 [14.82, 16.18] & 15.54 [14.87, 16.23] & 17.18 [16.43, 17.94] \\
I & Fog & M & 15.12 [14.47, 15.80] & 15.14 [14.49, 15.83] & 17.15 [16.41, 17.91] \\
I & Fog & H & 14.71 [14.07, 15.40] & 14.72 [14.08, 15.41] & 17.12 [16.38, 17.87] \\
I & Frame & E & 16.15 [15.45, 16.86] & 16.13 [15.43, 16.84] & 17.16 [16.42, 17.91] \\
I & Frame & M & 14.22 [13.56, 14.89] & 14.19 [13.53, 14.86] & 16.80 [16.07, 17.55] \\
I & Frame & H & 13.01 [12.37, 13.67] & 13.04 [12.40, 13.70] & 16.64 [15.92, 17.39] \\
I & Dark & E & 14.60 [13.93, 15.29] & 14.74 [14.06, 15.42] & 17.11 [16.38, 17.86] \\
I & Dark & M & 13.71 [13.05, 14.36] & 13.81 [13.16, 14.47] & 17.04 [16.31, 17.80] \\
I & Dark & H & 12.38 [11.79, 13.00] & 12.50 [11.90, 13.13] & 16.91 [16.18, 17.66] \\
I & Motion & E & 16.02 [15.32, 16.73] & 16.08 [15.37, 16.79] & 17.15 [16.41, 17.90] \\
I & Motion & M & 14.00 [13.34, 14.68] & 14.09 [13.43, 14.77] & 16.92 [16.19, 17.67] \\
I & Motion & H & 13.17 [12.52, 13.84] & 13.23 [12.57, 13.89] & 16.76 [16.04, 17.50] \\
I & Snow & E & 13.71 [13.01, 14.41] & 13.79 [13.08, 14.48] & 16.98 [16.23, 17.73] \\
I & Snow & M & 12.40 [11.77, 13.05] & 12.50 [11.85, 13.14] & 16.72 [15.99, 17.48] \\
I & Snow & H & 11.91 [11.29, 12.52] & 12.01 [11.40, 12.62] & 16.59 [15.86, 17.33] \\
P & Beam & E & 20.28 [19.29, 21.28] & 20.25 [19.27, 21.26] & 21.39 [20.39, 22.42] \\
P & Beam & M & 18.73 [17.84, 19.65] & 18.68 [17.79, 19.60] & 21.30 [20.30, 22.33] \\
P & Beam & H & 16.46 [15.62, 17.30] & 16.40 [15.57, 17.24] & 21.13 [20.15, 22.14] \\
P & Sensor & E & 18.21 [17.27, 19.17] & 18.19 [17.25, 19.15] & 21.32 [20.32, 22.34] \\
P & Sensor & M & 15.39 [14.59, 16.23] & 15.37 [14.56, 16.21] & 21.13 [20.14, 22.14] \\
P & Sensor & H & 13.84 [13.02, 14.66] & 13.81 [12.99, 14.63] & 20.99 [20.01, 22.00] \\
P & Cross & E & 20.90 [19.93, 21.89] & 20.89 [19.91, 21.88] & 21.43 [20.43, 22.46] \\
P & Cross & M & 20.17 [19.21, 21.14] & 20.15 [19.20, 21.13] & 21.43 [20.42, 22.47] \\
P & Cross & H & 19.39 [18.47, 20.34] & 19.39 [18.47, 20.35] & 21.42 [20.41, 22.45] \\
P & Fog & E & 20.25 [19.28, 21.22] & 20.22 [19.25, 21.19] & 21.32 [20.32, 22.36] \\
P & Fog & M & 16.60 [15.70, 17.49] & 16.53 [15.63, 17.42] & 21.05 [20.07, 22.07] \\
P & Fog & H & 6.48 [5.98, 6.95] & 6.46 [5.94, 6.94] & 20.45 [19.48, 21.43] \\
P & Echo & E & 20.51 [19.51, 21.54] & 20.50 [19.50, 21.54] & 21.41 [20.40, 22.44] \\
P & Echo & M & 19.94 [18.97, 20.95] & 19.93 [18.96, 20.95] & 21.37 [20.36, 22.40] \\
P & Echo & H & 18.48 [17.63, 19.36] & 18.45 [17.60, 19.33] & 21.24 [20.24, 22.27] \\
P & Motion & E & 18.93 [18.07, 19.81] & 18.91 [18.05, 19.79] & 21.39 [20.39, 22.42] \\
P & Motion & M & 17.04 [16.29, 17.80] & 17.01 [16.25, 17.76] & 21.30 [20.31, 22.32] \\
P & Motion & H & 15.42 [14.76, 16.08] & 15.36 [14.71, 16.01] & 21.22 [20.23, 22.22] \\
P & Snow & E & 20.94 [19.96, 21.94] & 20.94 [19.96, 21.94] & 21.44 [20.43, 22.47] \\
P & Snow & M & 20.42 [19.47, 21.41] & 20.42 [19.46, 21.40] & 21.44 [20.43, 22.47] \\
P & Snow & H & 19.72 [18.78, 20.70] & 19.72 [18.76, 20.70] & 21.41 [20.40, 22.44] \\
P & Wet & E & 21.12 [20.13, 22.14] & 21.12 [20.13, 22.13] & 21.43 [20.43, 22.46] \\
P & Wet & M & 20.93 [19.96, 21.94] & 20.93 [19.96, 21.94] & 21.43 [20.43, 22.46] \\
P & Wet & H & 20.53 [19.56, 21.53] & 20.55 [19.57, 21.55] & 21.43 [20.44, 22.46] \\
O & Dropout & E & 38.92 [37.84, 39.98] & 38.90 [37.82, 39.97] & 39.74 [38.62, 40.83] \\
O & Dropout & M & 34.09 [33.06, 35.10] & 33.95 [32.92, 34.96] & 39.46 [38.32, 40.56] \\
O & Dropout & H & 22.63 [21.53, 23.86] & 22.41 [21.30, 23.65] & 38.54 [37.36, 39.68] \\
O & Hole & E & 39.33 [38.19, 40.45] & 39.31 [38.17, 40.43] & 39.76 [38.64, 40.86] \\
O & Hole & M & 39.12 [37.99, 40.24] & 39.09 [37.96, 40.22] & 39.77 [38.64, 40.87] \\
O & Hole & H & 38.96 [37.82, 40.08] & 38.94 [37.80, 40.06] & 39.77 [38.64, 40.87] \\
O & Misalignment & E & 39.00 [37.89, 40.07] & 39.00 [37.89, 40.07] & 39.78 [38.66, 40.87] \\
O & Misalignment & M & 37.71 [36.59, 38.77] & 37.71 [36.59, 38.77] & 39.78 [38.66, 40.87] \\
O & Misalignment & H & 36.57 [35.46, 37.64] & 36.57 [35.46, 37.64] & 39.78 [38.66, 40.87] \\
O & Semantic & E & 33.83 [32.63, 34.83] & 33.83 [32.64, 34.83] & 39.75 [38.64, 40.84] \\
O & Semantic & M & 29.30 [28.36, 30.18] & 29.28 [28.34, 30.16] & 39.70 [38.58, 40.79] \\
O & Semantic & H & 25.42 [24.66, 26.17] & 25.40 [24.64, 26.16] & 39.56 [38.44, 40.64] \\
\midrule
O & Traffic & Fixed & \multicolumn{3}{c}{38.07 [37.01, 39.08]} \\
\end{longtable}
\endgroup
\clearpage

\subsection{External Validity of Manual Corruptions}
\label{sec:app_manual_external_validity}

We compare the manual interventions with residuals from clean STCOcc and
SDGOcc predictions on all 6,019 ordered frames from the 150 Occ3D-nuScenes
validation scenes. Fifty scenes are used only to fix the 5th--95th percentile
comparison envelopes of component-level morphology descriptors, and the
remaining 100 scenes are held out for evaluation. The descriptors measure
component volume, bounding-box volume, fill ratio, ego range, and dynamic or
interior fractions where applicable. All comparisons use the common Occ3D
\texttt{mask\_camera} domain. Components smaller than four voxels are excluded,
and at most 128 components of each error type are retained per frame. This
retains at least 68.49\% of the residual voxel mass in every reported source
and error-type pair. Coverage requires all applicable descriptors to lie within
their calibration envelopes. Component and voxel coverage are the unweighted
and voxel-weighted fractions of retained components satisfying this condition.
The confidence intervals in
Table~\ref{tab:app_external_coverage} use 10,000 scene-clustered bootstrap
replicates with seed 20260726.

\begin{table}[H]
    \centering
    \scriptsize
    \setlength{\tabcolsep}{4pt}
    \renewcommand{\arraystretch}{1.08}
    \caption{Held-out morphology coverage of clean-input upstream residuals.
    Coverage and 95\% confidence intervals are percentages. Components are
    retained after the filtering described in the text.}
    \label{tab:app_external_coverage}
    \begin{adjustbox}{max width=\linewidth}
    \begin{tabular}{@{}llrcc@{}}
        \toprule
        \textbf{Manual family} & \textbf{Source} & \textbf{Components} &
        \textbf{Component coverage} & \textbf{Voxel coverage} \\
        \midrule
        Semantic & STCOcc & 202,173 & 74.6 [73.5, 75.7] & 74.8 [71.4, 77.9] \\
                 & SDGOcc & 309,402 & 71.1 [70.1, 72.0] & 68.0 [65.4, 70.6] \\
        Dropout  & STCOcc & 507,584 & 71.5 [70.3, 72.6] & 44.2 [42.1, 46.3] \\
                 & SDGOcc & 295,502 & 69.1 [67.5, 70.7] & 60.3 [56.2, 64.2] \\
        Hole     & STCOcc & 82      & 78.0 [59.2, 93.7] & 18.1 [6.2, 64.5] \\
                 & SDGOcc & 355     & 83.7 [76.7, 90.2] & 41.7 [29.0, 65.2] \\
        \bottomrule
    \end{tabular}
    \end{adjustbox}
\end{table}

The coverage results support overlap in spatial morphology, not equality with
a deployment failure distribution. Semantic replacement covers substantial
natural error support, but its class-transition Jensen--Shannon divergence is
0.72--0.89 bits and its morphology-based classifier two-sample test (C2ST)
AUC is 0.80--0.82. Its spatial
footprint is therefore relevant, while the replacement probabilities are not
an empirical confusion distribution. Dropout has the strongest defensible
correspondence to false-negative residuals, with C2ST AUC 0.638--0.664 for the
best-matched moderate and hard settings, although its severity remains
source-dependent. A C2ST AUC near 0.5 means that the compared descriptor
distributions are difficult to distinguish, while higher values indicate a
mismatch. Dynamic-interior false negatives used to assess hole occur
in only 1.9\% of held-out STCOcc frames and 7.8\% of SDGOcc frames, so hole is
a rare topology diagnostic. Correctly calibrated clean predictions cannot
establish the prevalence of pose failures. The clean-output translation proxy
places the manual 0.644--1.733 m shifts at the 99.3rd--100th percentiles, so
misalignment remains a controlled calibration stressor. Traffic is excluded
from this residual comparison because it is a coordinate-consistent regional
layout transfer applied to both inputs and targets.

We next replay semantic-only and false-negative-only residuals into otherwise
clean I$^2$-World H4 inputs. Manual severity is fixed before inference by
selecting the level whose held-out residual-burden percentile is nearest 50.
The replay uses the 4,519 strict H4/F6 anchors from all 150 scenes and six
future horizons. We use 10,000 paired scene-clustered bootstrap replicates with
seed 20260727. The 6,019-frame morphology audit and the 4,519-anchor replay serve different
purposes and do not mix evaluation populations. Table~\ref{tab:app_residual_replay}
reports the 1s, 2s, and 3s-averaged mIoU drops and the cosine similarity between the
six-horizon drop vectors.

\begin{table}[H]
    \centering
    \scriptsize
    \setlength{\tabcolsep}{3pt}
    \renewcommand{\arraystretch}{1.05}
    \caption{I$^2$-World response to clean-input upstream residuals and matched
    manual interventions. The norm ratio is the magnitude of the natural
    six-horizon drop vector divided by that of its manual counterpart.}
    \label{tab:app_residual_replay}
    \begin{adjustbox}{max width=\linewidth}
    \begin{tabular}{@{}llllrrrr@{}}
        \toprule
        \textbf{Source} & \textbf{Residual} & \textbf{Manual match} &
        \textbf{Protocol} & \textbf{Natural drop} & \textbf{Manual drop} &
        \textbf{Cosine [95\% CI]} & \textbf{Norm ratio} \\
        \midrule
        STCOcc & Semantic & Semantic/mid  & \curprot{}   & 2.778 & 10.480 & 0.999 [0.997, 1.000] & 0.274 \\
               &          &               & \burstprot{} & 2.791 & 10.502 & 0.999 [0.997, 1.000] & 0.275 \\
               &          &               & \histprot{}  & 0.051 & 0.078  & 0.956 [0.632, 0.986] & 0.489 \\
        SDGOcc & Semantic & Semantic/hard & \curprot{}   & 4.690 & 14.360 & 1.000 [0.999, 1.000] & 0.333 \\
               &          &               & \burstprot{} & 4.710 & 14.376 & 1.000 [0.999, 1.000] & 0.334 \\
               &          &               & \histprot{}  & 0.049 & 0.221  & 0.964 [0.612, 0.992] & 0.199 \\
        STCOcc & False negative & Dropout/hard & \curprot{}   & 8.828 & 17.144 & 1.000 [0.999, 1.000] & 0.514 \\
               &                &              & \burstprot{} & 8.871 & 17.363 & 1.000 [0.999, 1.000] & 0.510 \\
               &                &              & \histprot{}  & 0.183 & 1.234  & 0.986 [0.936, 0.997] & 0.174 \\
        SDGOcc & False negative & Dropout/mid  & \curprot{}   & 1.240 & 5.688 & 0.992 [0.985, 0.996] & 0.238 \\
               &                &              & \burstprot{} & 1.269 & 5.828 & 0.992 [0.986, 0.996] & 0.237 \\
               &                &              & \histprot{}  & 0.064 & 0.321 & 0.975 [0.859, 0.994] & 0.178 \\
        \bottomrule
    \end{tabular}
    \end{adjustbox}
\end{table}

Across the 12 comparisons, cosine similarity is 0.956--1.000 with a median of
0.996. The median for \textit{Current-only} and \textit{Recent-burst} is
0.999. Natural residuals have only 17--51\% of the matched manual response
magnitude. The manual interventions closely match the direction and horizon
profile under \textit{Current-only} and \textit{Recent-burst}, with larger
degradation magnitudes. \textit{History-only} effects are small and their
profile agreement is less certain. Natural high-error episodes have median durations of
2--4 frames across the audited error types, which supports the relevance of
the three temporal protocols but not the full long-duration tail. Finally,
false-positive occupancy accounts for 18.5\% and 34.7\% of STCOcc and SDGOcc
residual voxels, respectively, and has no direct manual counterpart. These
results support construct and downstream-response validity conditional on the
evaluated dataset, upstream sources, and forecaster. They do not estimate
fleet-level failure prevalence or deployment risk.

\subsection{Manual Corruption Realization Audit}

We separately regenerate the stochastic manual corruptions using five
independent realizations: the released deterministic realization and generator
seeds 1 through 4. Semantic replacement, dropout, and hole are evaluated at
hard severity over the full 5,119-frame manual content inventory. Table~\ref{tab:app_multiseed}
reports variation in aggregate changed-voxel magnitude and spatial overlap.

\begin{table}[H]
    \centering
    \small
    \setlength{\tabcolsep}{6pt}
    \renewcommand{\arraystretch}{1.10}
    \caption{Five-realization sensitivity of stochastic manual corruptions.
    CV is computed across realization-level mean changed-voxel counts.
    Jaccard pools full-grid changed-voxel intersections and unions over all
    5,119 frames for each pair, then averages the ten realization pairs.}
    \label{tab:app_multiseed}
    \begin{tabular}{@{}lccc@{}}
        \toprule
        \textbf{Family} & \textbf{Changed voxels} & \textbf{CV} &
        \textbf{Pairwise Jaccard} \\
        \midrule
        Semantic & 2,939.6 & 0.69\% & 0.310 \\
        Dropout & 12,996.6 & 0.19\% & 0.447 \\
        Hole & 246.7 & 2.24\% & 0.868 \\
        \bottomrule
    \end{tabular}
\end{table}

The low between-realization CVs show that aggregate corruption magnitude is
stable. Semantic and dropout nevertheless produce spatially distinct
realizations, while hole is more constrained by eligible object interiors.
This audit measures generator-level magnitude and spatial variation rather
than downstream model-score variation.

\section{Additional Diagnostic Visualizations}

This section collects the supplementary visual diagnostics used to interpret
the aggregate OccStress tables. The figures are grouped by the point in the
forecasting chain that they inspect: upstream source quality, controlled manual
state corruptions, traffic-layout transfer, and qualitative predictions under
different temporal protocols.

\subsection{Source and Upstream Diagnostics}

\begin{figure}[H]
    \centering
    \includegraphics[width=\textwidth]{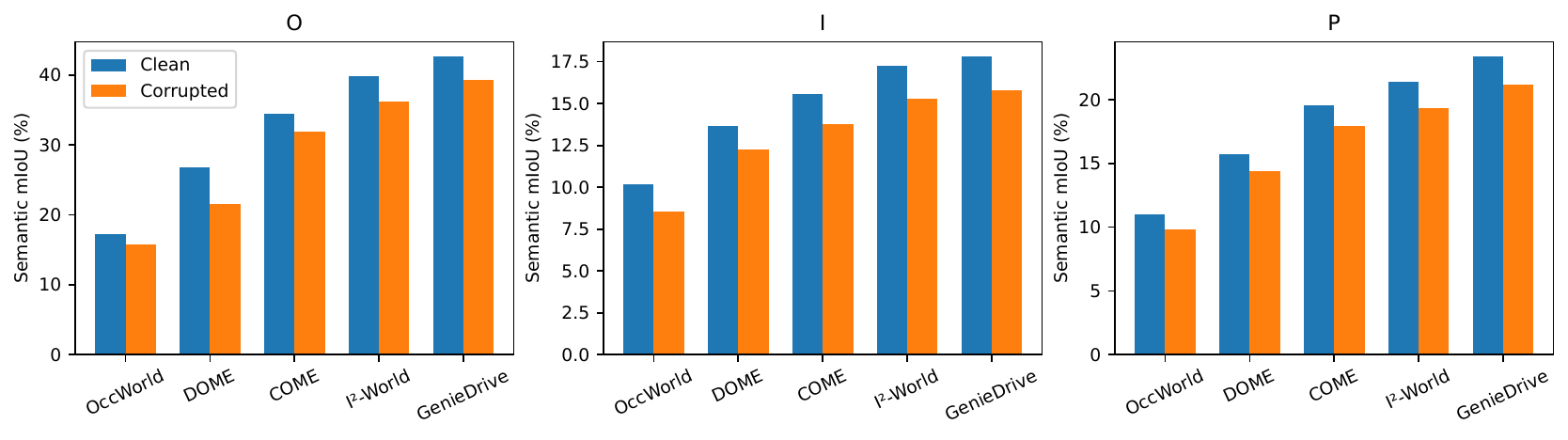}
    \caption{Clean source quality and average robustness across the manual,
    camera-only, and point-cloud-fusion tracks. The figure separates downstream
    forecasting robustness from the clean-source reference performance.}
    \label{fig:app_source_bottleneck}
\end{figure}

Figure~\ref{fig:app_source_bottleneck} separates the quality of the clean
state source from robustness under corruption. It provides a visual counterpart
to the \(UCR\), \(CS^S\), and \(RS^S_{\forall}\) quantities used in
Table~\ref{tab:4_occstress_summary}.

\begin{figure}[H]
    \centering
    \includegraphics[width=0.98\textwidth]{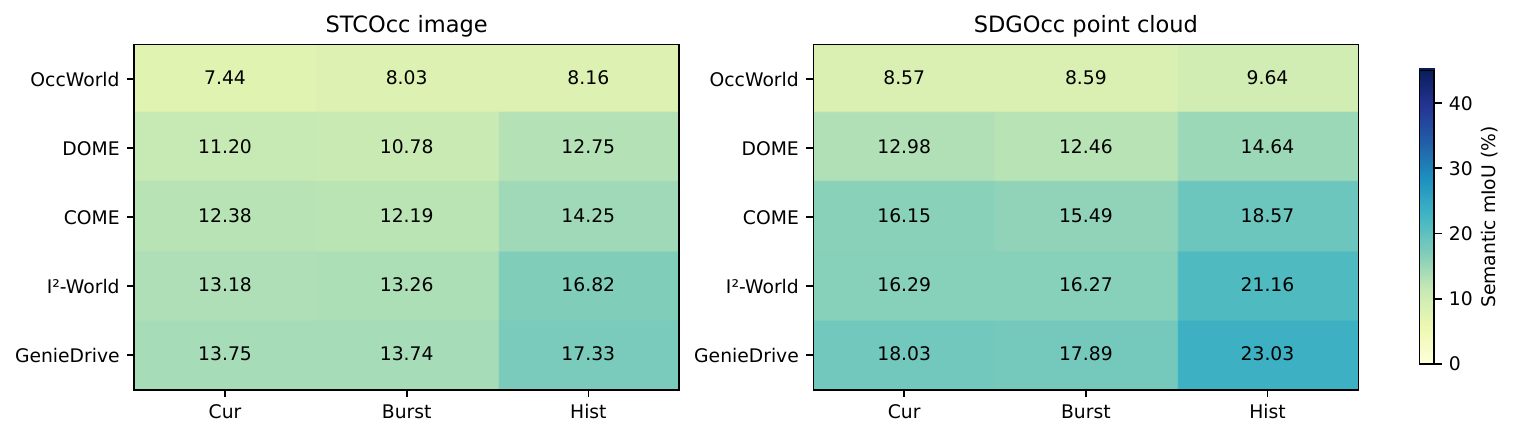}
    \caption{Upstream 3D occupancy corruption propagated into 4D forecasting.
    Camera-only panels use STCOcc hard corruptions; point-cloud-fusion panels
    use SDGOcc heavy corruptions. Each cell reports horizon-averaged semantic
    mIoU under the corresponding temporal protocol.}
    \label{fig:app_upstream_protocol_heatmaps}
\end{figure}

\begin{figure}[H]
    \centering
    \includegraphics[width=0.98\textwidth]{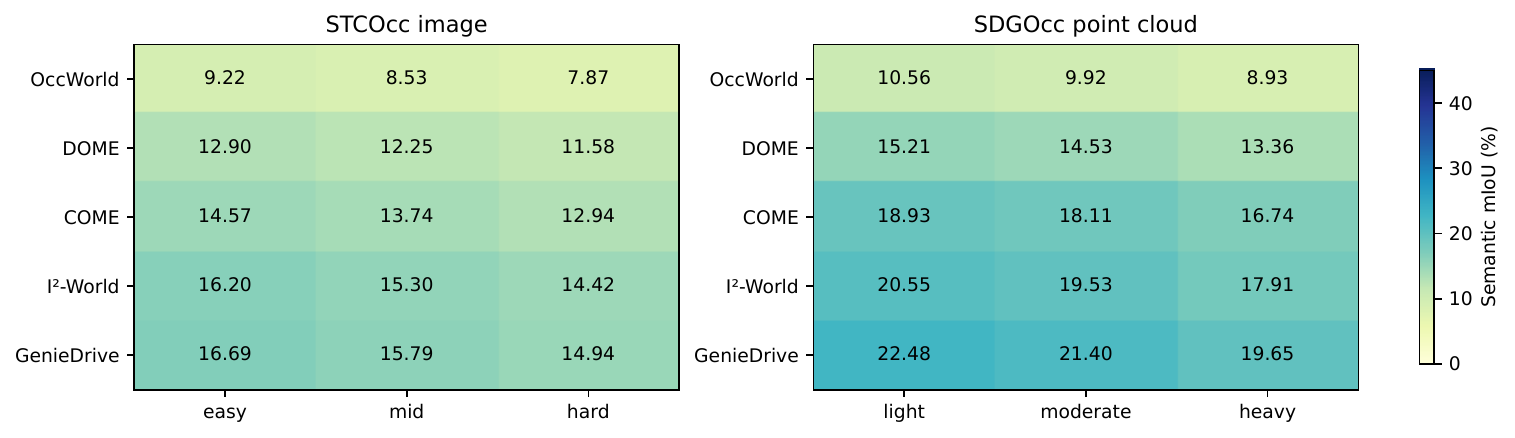}
    \caption{Upstream corruption severity sweeps averaged over temporal
    protocols. Camera-only columns correspond to easy, mid, and hard
    severities; point-cloud-fusion columns correspond to light, moderate, and
    heavy severities.}
    \label{fig:app_upstream_severity_heatmaps}
\end{figure}

Figures~\ref{fig:app_upstream_protocol_heatmaps} and
\ref{fig:app_upstream_severity_heatmaps} provide a more detailed view of the
upstream track. The protocol heatmaps show how corrupted upstream occupancy
states interact with temporal placement, while the severity heatmaps show which
observation corruptions remain harmful after the 3D occupancy source has
processed the sensor input.

\subsection{Manual State-Corruption Diagnostics}

\begin{figure}[H]
    \centering
    \includegraphics[width=\textwidth,height=0.82\textheight,keepaspectratio]{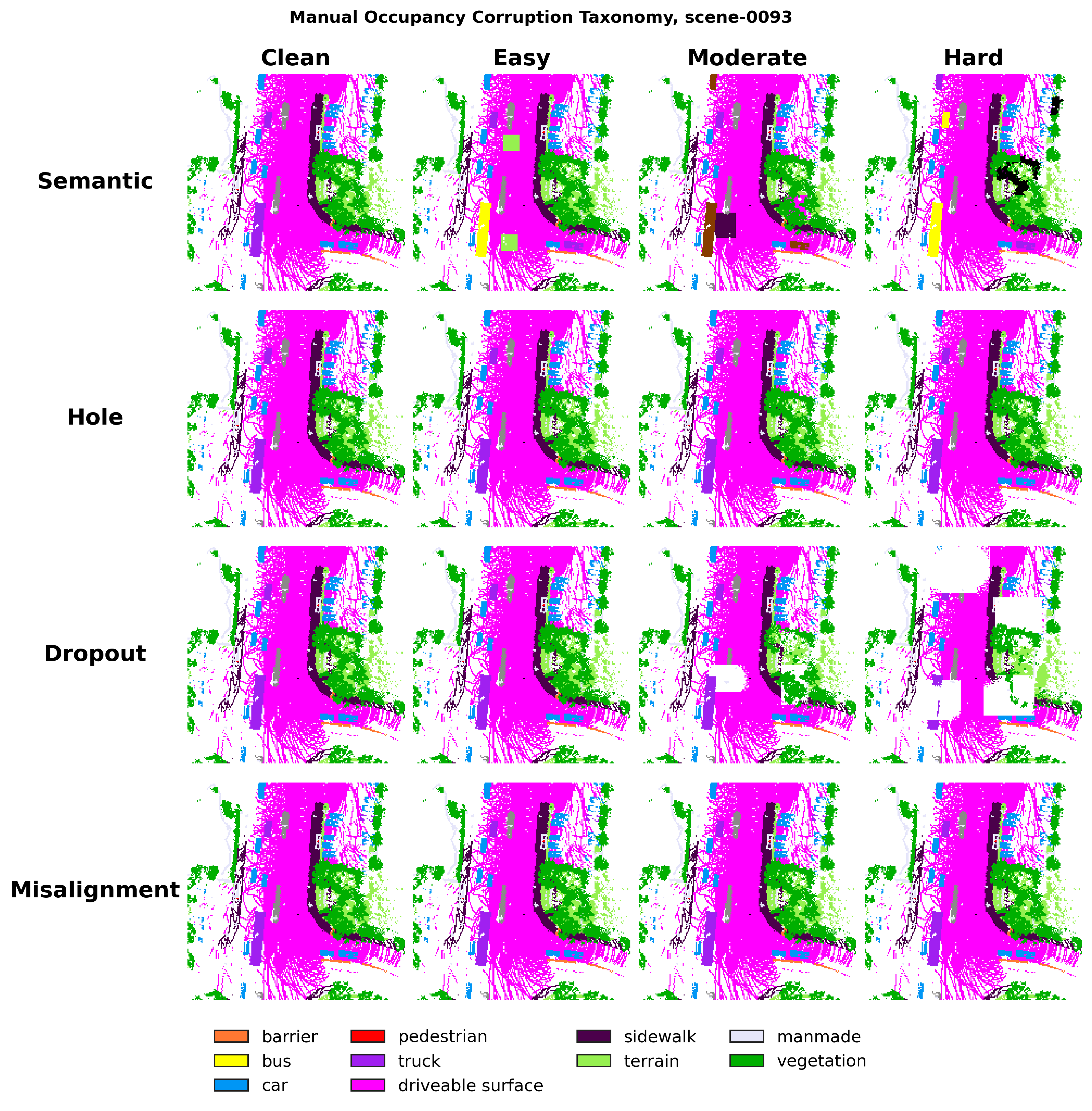}
    \caption{BEV visualization of manual occupancy-state corruptions on a
    shared scene. Each row compares the clean input with easy, moderate, and
    hard variants using the same crop, projection, and Occ3D semantic color
    map.}
    \label{fig:app_manual_taxonomy}
\end{figure}

Figure~\ref{fig:app_manual_taxonomy} visualizes the manual corruption families
before they enter the 4D forecaster. Traffic is excluded from this taxonomy
figure because it is a coordinate-consistent mirrored-scene transformation
rather than a severity ladder applied to a fixed target sequence.

\begin{figure}[H]
    \centering
    \begin{subfigure}{0.98\textwidth}
        \centering
        \includegraphics[width=\linewidth]{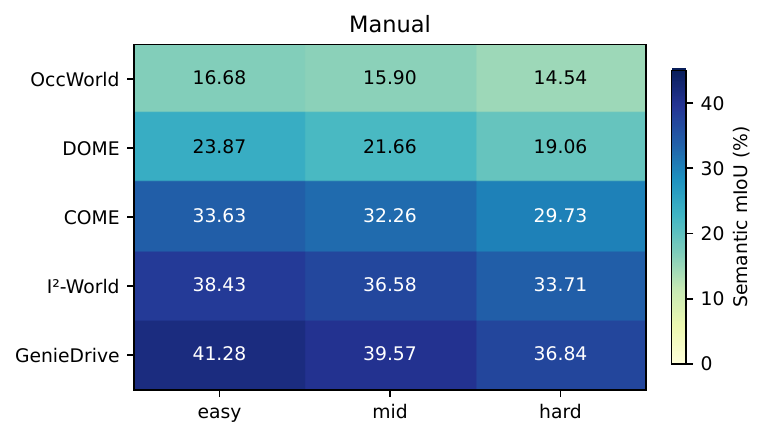}
        \caption{Severity sweep averaged over temporal protocols.}
    \end{subfigure}
    \vspace{0.6em}
    \begin{subfigure}{0.98\textwidth}
        \centering
        \includegraphics[width=\linewidth]{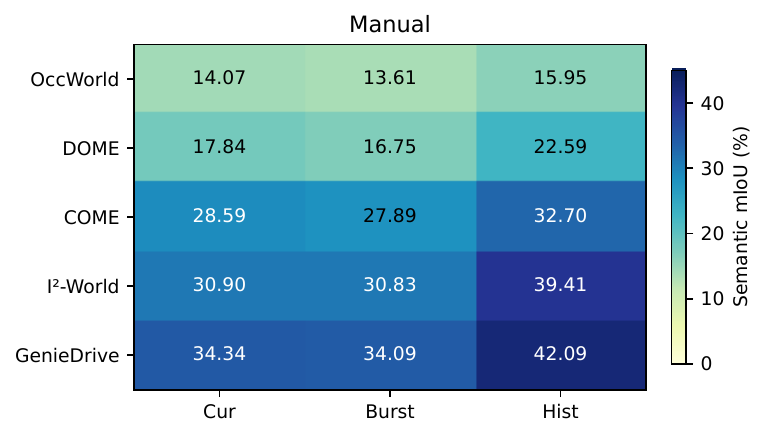}
        \caption{Hard corruptions under \textit{Current-only}, \textit{Recent-burst}, and
        \textit{History-only} protocols.}
    \end{subfigure}
    \caption{Manual occupancy-state corruption diagnostics. Each heatmap cell
    reports horizon-averaged semantic mIoU; higher is better.}
    \label{fig:app_manual_heatmaps}
\end{figure}

Figure~\ref{fig:app_manual_heatmaps} complements
Table~\ref{tab:app_manual_protocol} by separating manual corruptions along two
axes: severity and temporal protocol. Both panels average the four non-traffic families. The severity panel summarizes degradation as corruption strength increases, while the protocol panel separates temporal placement.

\begin{figure}[H]
    \centering
    \begin{subfigure}{0.98\textwidth}
        \centering
        \includegraphics[width=\linewidth]{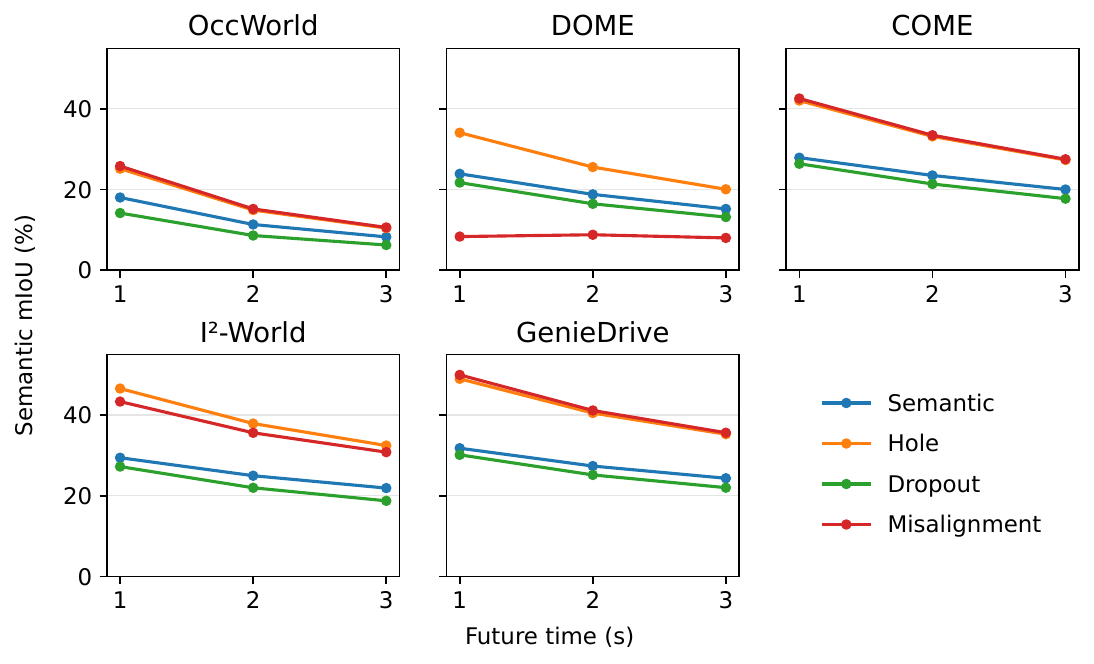}
        \caption{Horizon-wise mIoU under \textit{current-only} hard manual corruption.}
    \end{subfigure}
    \vspace{0.6em}
    \begin{subfigure}{0.98\textwidth}
        \centering
        \includegraphics[width=\linewidth]{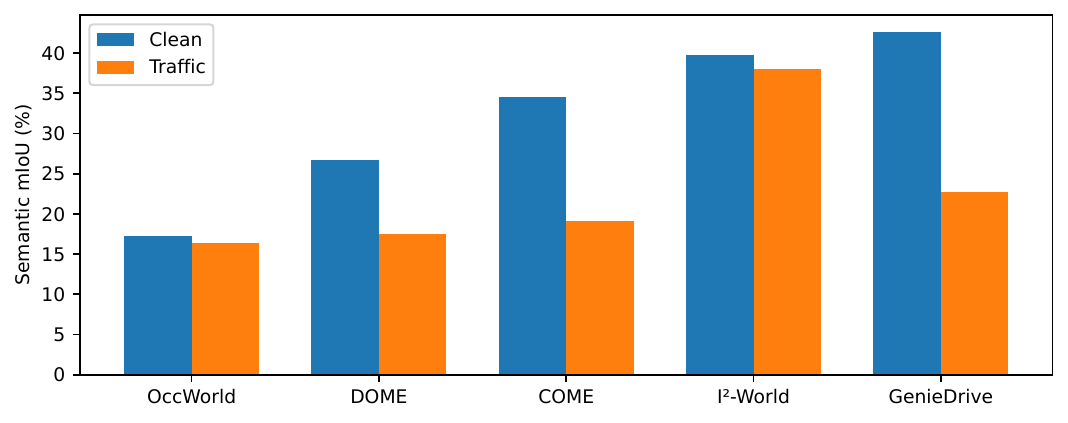}
        \caption{Traffic mirroring compared with clean evaluation.}
    \end{subfigure}
    \caption{Manual-track horizon and traffic diagnostics. Traffic is a
    structured scene transformation and is therefore shown separately from
    noisy-input state corruptions.}
    \label{fig:app_manual_horizon_traffic}
\end{figure}

Figure~\ref{fig:app_manual_horizon_traffic} shows two diagnostics that are
reported separately from the main manual corruption averages. The horizon
curves expose how \textit{current-only} hard manual corruptions evolve over forecasting
horizons. The traffic panel compares clean evaluation with the valid
coordinate-consistent traffic mirroring protocol.

\subsection{Qualitative Prediction Grids}

\begin{figure}[H]
    \centering
    \includegraphics[width=\textwidth]{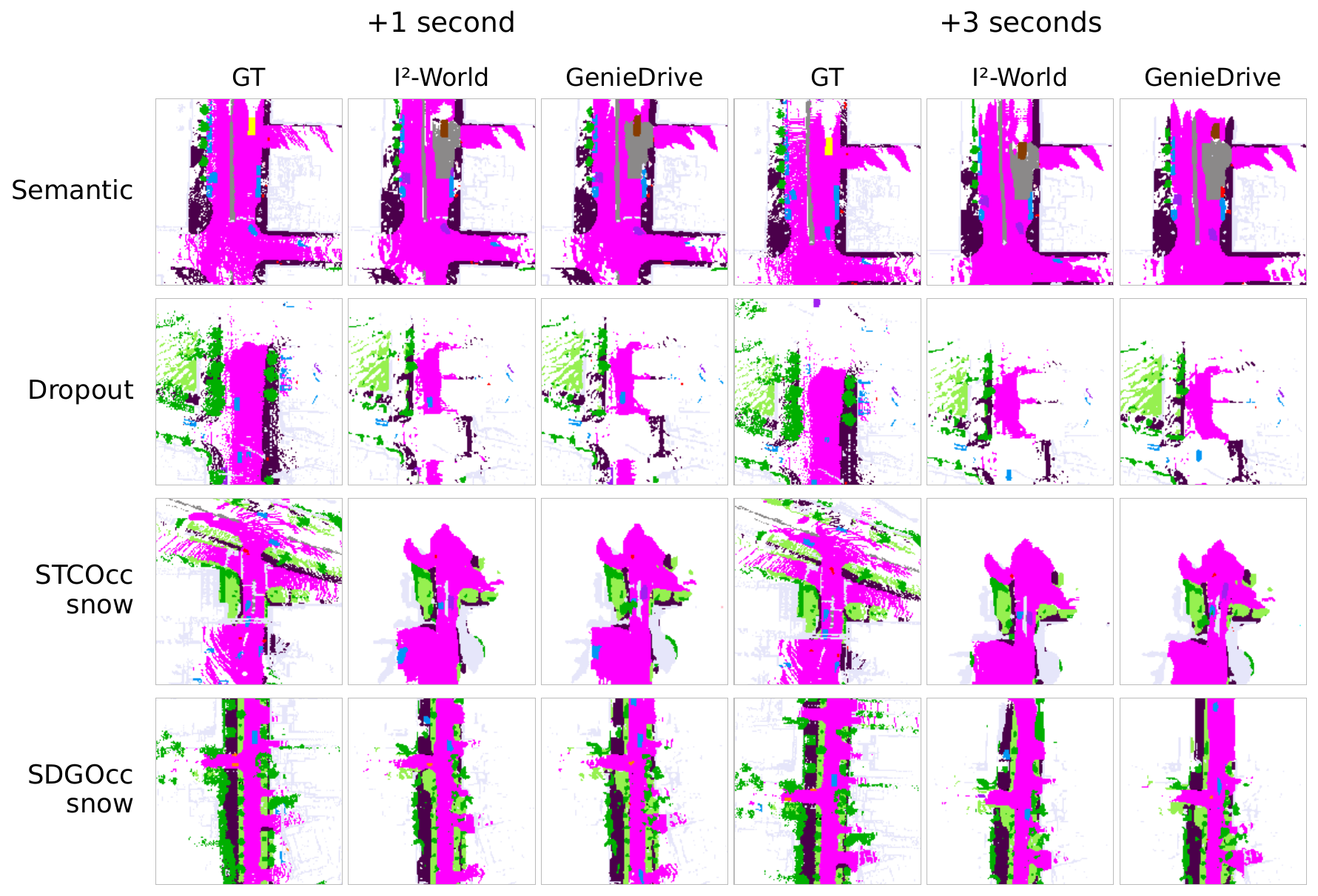}
    \caption{Matched I$^2$-World and GenieDrive predictions under the
    \textit{Current-only} hard protocol. Each row uses one shared scene and anchor.}
    \label{fig:app_i2world_grid_current}
\end{figure}

\begin{figure}[H]
    \centering
    \includegraphics[width=\textwidth]{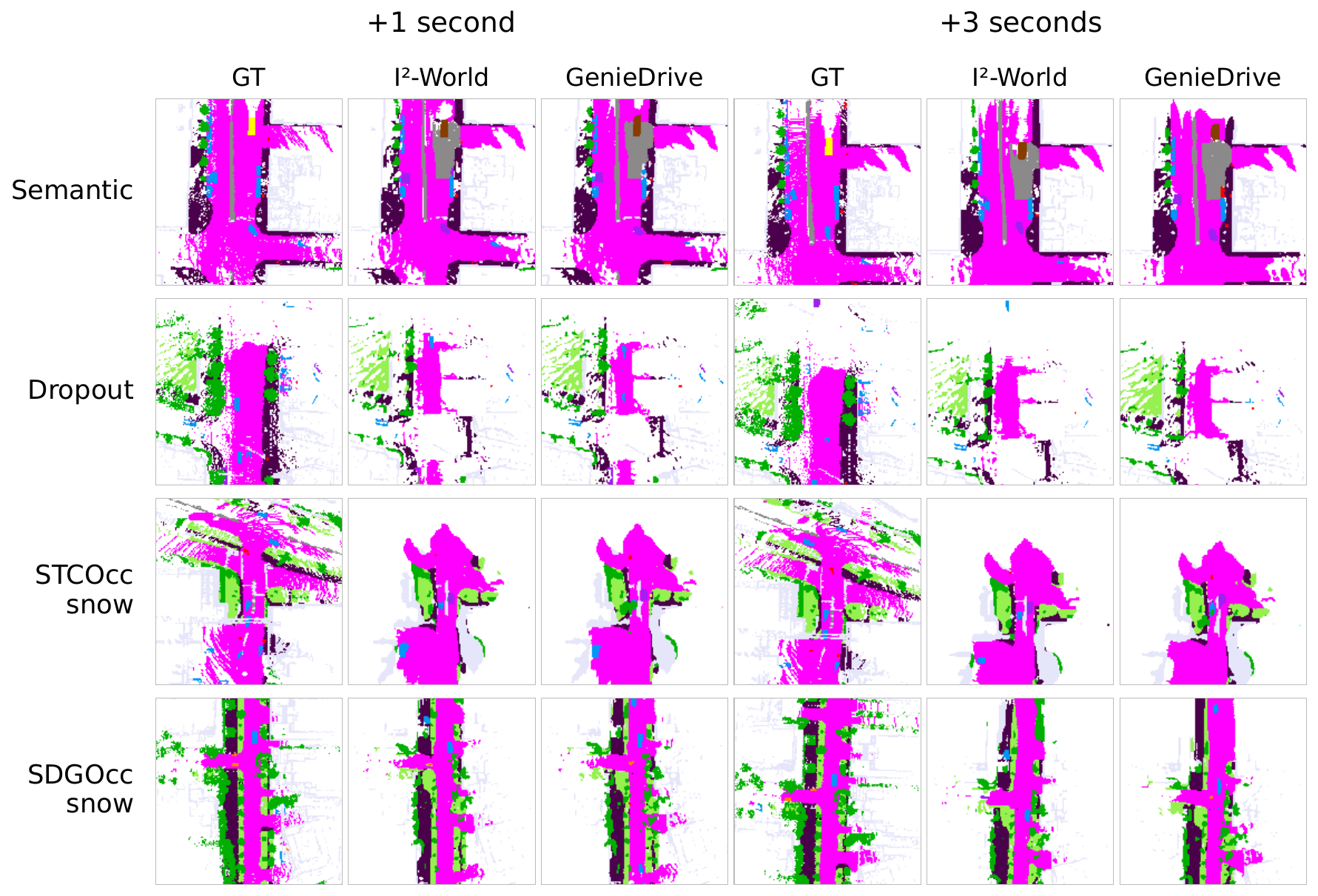}
    \caption{Matched I$^2$-World and GenieDrive predictions under the
    \textit{Recent-burst} hard protocol, using the same anchors as
    Figure~\ref{fig:app_i2world_grid_current}.}
    \label{fig:app_i2world_grid_burst}
\end{figure}

\begin{figure}[H]
    \centering
    \includegraphics[width=\textwidth]{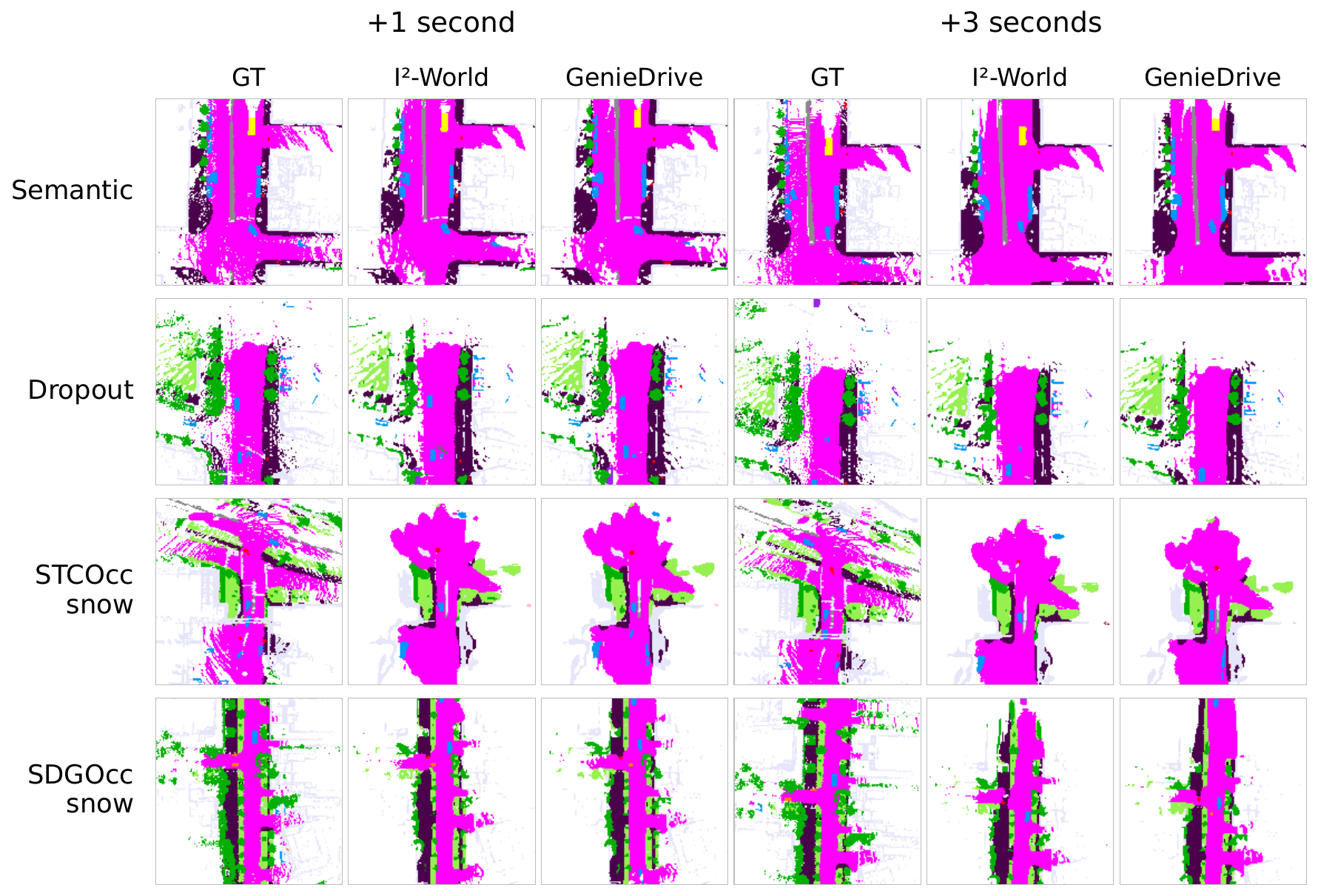}
    \caption{Matched I$^2$-World and GenieDrive predictions under the
    \textit{History-only} hard protocol, using the same anchors as
    Figure~\ref{fig:app_i2world_grid_current}.}
    \label{fig:app_i2world_grid_history}
\end{figure}

Figures~\ref{fig:app_i2world_grid_current}--\ref{fig:app_i2world_grid_history}
compare I$^2$-World and GenieDrive under the three temporal protocols.
Each row uses the same anchor for both models and across protocols, selected
as a representative case near the median scene and corruption features,
without using forecasting scores. Rows cover manual semantic and dropout
corruptions, STCOcc image snow, and SDGOcc point-cloud snow at hard severity.
The two column groups show ground truth and both models' predictions at 1s
and 3s, using the same full-grid BEV projection and semantic colors.

\clearpage
\flushbottom

\end{document}